\documentclass[num-refs, twocolumn]{wiley-article}

\usepackage{siunitx}

\usepackage{booktabs}
\usepackage{multirow}
\usepackage[justification=justified]{caption}
\usepackage{subcaption}
\usepackage{float}
\usepackage{graphicx}
\RequirePackage[hidelinks]{hyperref}

\title{A Coarse to Fine 3D LiDAR Localization with Deep Local Features for Long Term Robot Navigation in Large Environments}

\author[1]{Míriam Máximo} \author[1]{Antonio Santo} \author[1]{Arturo Gil} \author[1]{Mónica Ballesta} \author[1]{David Valiente}

\affil[1]{University Institute for Engineering Research, Miguel Hernández University, Avda. de la Universidad s/n, 03202 Elche, Alicante, Spain}

\corraddress{Corresponding author Míriam Máximo.}
\corremail{mmaximo@umh.es}

\runningauthor{M. Máximo et al.}

\begin{document}

\begin{frontmatter}
\maketitle

\begin{abstract}
The location of a robot is a key aspect in the field of mobile robotics. This problem is particularly complex when the initial pose of the robot is unknown. In order to find a solution, it is necessary to perform a global localization. In this paper, we propose a method that addresses this problem using a coarse-to-fine solution. The coarse localization relies on a probabilistic approach of the Monte Carlo Localization (MCL) method, with the contribution of a robust deep learning model, the MinkUNeXt neural network, to produce a robust description of point clouds of a 3D LiDAR within the observation model. The MCL method has been approached from a topological perspective, considering that the particles are initialized on the map positions where LiDAR scans have been previously captured. For fine localization, global point cloud registration has been implemented. MinkUNeXt aids this by exploiting the outputs of its intermediate layers to produce deep local features for each point in a scan. These features facilitate precise alignment between the current sensor observation (query) and one of the point clouds on the map. The proposed MCL method incorporating Deep Local Features for fine localization is termed MCL-DLF. Alternatively, a classical ICP method has been implemented for this precise localization aiming at comparison purposes. This method is termed MCL-ICP. In order to validate the performance of MCL-DLF method, it has been tested on publicly available datasets such as the NCLT dataset, which provides seasonal large-scale environments. Additionally, tests have been also performed with own data (UMH) that also includes seasonal variations on large indoor/outdoor scenarios. The results, which were compared with established state-of-the-art methodologies, demonstrate that the MCL-DLF method obtains an accurate estimate of the robot localization in dynamic environments despite changes in environmental conditions. For reproducibility purposes, the code is publicly available \footnotemark.

\end{abstract}

\end{frontmatter}

\section{Introduction}\label{introduction}
\footnotetext{\url{https://github.com/miriammaximo/MCL-DLF.git}}
The fundamental characteristic of a mobile robot \cite{ullah2024mobile} lies in its capacity for autonomous navigation. This capability needs precise determination of its position and orientation, a problem commonly referred to as localization \cite{kramer2012accurate}. The algorithms developed to address this localization challenge can be broadly categorized as global or local, depending on the robot’s initial state.

Global localization algorithms are designed to operate under the assumption that the robot’s initial pose is unknown, a scenario often termed the kidnapped robot problem. These methods enable the coarse estimation of the robot’s pose leveraging sensor measurements. 

However, some of these sensors have limitations. For instance, Global Navigation Satellite System (GNSS)-based positioning systems do not provide reliable data. This is due to the presence of significant errors, particularly in areas with close proximity to buildings, where there is no reliable satellite signal. Furthermore, this signal is not available in indoor environments either, rendering these sensors unsuitable for such applications and necessitating alternative sensing modalities for accurate localization. Other sensors that provide external environmental information include cameras, radar, and LiDAR.

The use of cameras \cite{heredia2025static} enables a localization system based on visual perception of the environment. Image features are extracted through a variety of methods, ranging from more classical approaches \cite{dalal2005histograms, rublee2011orb} to more recent methods that integrate deep learning networks \cite{detone2018superpoint}. While cameras offer a rich source of environmental information and are cost-effective, they are highly susceptible to variations in lighting conditions and environmental factors such as fog or rain, limiting their applicability in dynamic environments.

In contrast, LiDAR (Light Detection and Ranging) sensors \cite{yin2024survey} exhibit superior robustness to changes in illumination and environmental conditions. By employing laser pulses to measure distances and construct 3D environmental maps, LiDAR sensors provide high accuracy in obstacle and environmental feature detection \cite{wang2022accurate, technologies13040162}.

Within the context of global localization, place recognition methods play a pivotal role \cite{fan2022svt, luo2023bevplace}. These methods aim to identify previously visited locations by comparing current sensor measurements against a database of known maps, frequently constructed using SLAM techniques \cite{zhang20243d}.  In many instances, this comparison is achieved through methods that infer the similarity between two point clouds, often by embedding each point cloud into a global descriptor and subsequently comparing these descriptors using techniques such Euclidean distance or cosine distance for nearest neighbor search.

A common technique for global localization is MCL \cite{dellaert1999monte}, a particle filter-based method. A key aspect of MCL is the utilization of environmental information to update the particle set, representing hypotheses of the robot’s pose. This information can be derived from various sources, including image features \cite{fernandez2012monte, xu2019robust}, LiDAR scan features \cite{zeng2023indoor}, or sensor fusion \cite{9476732}. Thus, incorporating point cloud similarity into the observation model allows for more robust localization compared to a simple nearest neighbor search.

Place recognition methods \cite{7339473} typically yield a coarse pose estimate, necessitating the integration of local localization methods for refined pose estimation. These local localization algorithms aim to determine the robot’s pose relative to its immediate surroundings. Dead reckoning, which estimates the robot’s current pose based on its last known position and motion measurements from sensors like wheel encoders, is a common approach. However, it is prone to error accumulation due to factors such slip or friction of the wheels. Thus, it is necessary to combine it with other methods to achieve greater accuracy.

Point cloud registration algorithms are employed for precise local localization. This can be addressed with the position of the points, as in Iterative Closest Point (ICP) \cite{besl1992method} and related methods \cite{9925083}, or by utilizing 3D point cloud features \cite{rusu2009fast, salti2014shot}. In some cases this features are obtained with a deep neural network \cite{wang2019deep, yew2020rpm}.

All in all, to implement a complete localization method, a coarse-to-fine approach is often adopted.  This strategy involves initially obtaining an approximate global pose, subsequently refined through local localization techniques. Some methods focus on extracting local features and global descriptors from point clouds using the same network \cite{9645340}. However, in certain instances, it is requisite for the architecture to comprise a dedicated network branch for the derivation of the global descriptor and a separate branch for the extraction of keypoints and local descriptors, thereby necessitating additional convolution operations for these outputs.

In this paper,  we explore a similar coarse-to-fine localization strategy. For coarse localization, we employ MCL with an observation model that compares point clouds. This comparison is performed by embedding the point clouds into a deep learning descriptor using the MinkUNeXt network \cite{cabrera2024minkunext}, a U-Net architecture that utilizes 3D sparse convolutions.

The implementation of sparse networks facilitates the processing of three-dimensional data, which can attain excessively large dimensions. Given the inherent sparsity of 3D point clouds, these networks efficiently process only occupied points or voxels, significantly reducing computational overhead compared to dense networks. Additionally, U-Net architectures \cite{ronneberger2015u}, with their encoder-decoder design, effectively capture spatial relationships, detailing 3D geometry analysis, focusing on relevant data while mitigating the impact of noise.

Following the coarse localization stage, a fine localization method is applied. This method analyzes correspondences between local features of LiDAR scans, comparing the current scan with a map scan from the region identified by the coarse localization. Feature extraction is performed using the same network employed in coarse localization, but in this case utilizing intermediate layer outputs. Notably, notwithstanding that this network is not specifically trained to obtain local features. This paper aims to investigate point cloud registration using sparse networks without relying on predefined keypoints.

The proposed methodology enables precise robot pose estimation without prior knowledge of the initial location. This approach has been evaluated in environments characterized by dynamic changes, including the presence of people, vehicles, and seasonal variations.

The contributions of this paper are presented as follows:
\vspace{-0.2cm}
\begin{itemize}
    \item A complete method, termed MCL-DLF, for the coarse-to-fine localization of mobile robots in large scale environments and environmental changing scenarios. This approach is suitable for both outdoor and indoor environments, achieving robust performance during handover situations, specifically transitions from outdoor to indoor settings where environmental conditions undergo significant alterations.
    \item An implementation of the MCL method with an observation model that allows robust coarse localization by using point clouds embedded in a deep learning descriptor. This descriptor is provided by the use of a 3D sparse convolutional neural, MinkUNeXt.
    \item A point cloud registration method for fine localization that reduces position and orientation errors to achieve an accurate solution to the robot's pose. This method involves the correspondence between local features of the point cloud, obtained through the intermediate layers of the MinkUNeXt network. These correspondences are then used to perform the registration with RANSAC.
\end{itemize}

\section{Related work}\label{related_work}
\subsection{Point cloud-based localization}
Localization based on 3D LiDAR data has become very popular in recent years due to the fact that this type of sensor is able to capture the elements of the environment with a high degree of accuracy. Furthermore, its robustness to variations in illumination and seasonal conditions makes it preferable to other sensors such as cameras.

There are two approaches to LiDAR localization: coarse localization, when the robot's position is initially unknown; and point cloud registration,  which refines the position after it is approximately estimated.

\subsubsection{Coarse localization}
Several publications have focused on identifying previously visited locations, a task known as place recognition. This is accomplished using 3D data by comparing the similarity between point cloud representations to retrieve a query from a map database. Point clouds are encoded as descriptors that encapsulate the global geometric features of the scene, a process facilitated by deep learning methodologies.

One of the earliest studies related to this topic was PointNetVLAD \cite{uy2018pointnetvlad}. In this work, PointNet \cite{qi2017pointnet} and NetVLAD \cite{arandjelovic2016netvlad} are combined. The extraction of local features is facilitated by PointNet, and these features are then compactly grouped into a single descriptor by NetVLAD. The training is performed by applying metric learning, so that it is capable of generating similar descriptors for similar areas. This is done by applying a loss function called "lazy triplet and quadruplet" which maximizes the differences between descriptors of distant point clouds.

Further work \cite{schaupp2019oreos} is also focused on using point clouds in place recognition, including the relative orientation between the two point clouds. In this case, the point cloud is converted to a 2D range image to be used as input to the network. The network consists of a convolutional neural network that provides two outputs: a rotation-invariant vector $v$ and a vector $w$ that embeds the rotation information. The vector $v$ is compared with the vectors of the map's point clouds to perform a nearest neighbor search, and once the closest one is obtained, the relative orientation is obtained using the $w$ vectors of each cloud.

On \cite{9359460}, it also focuses on the orientation between point clouds. The network input, in this case, is a 2D image obtained from the horizontal plane projection of a point cloud. Utilizing a feature extraction module, 2D images are transformed into their frequency domain representation, facilitating the creation of distinct place descriptors. Furthermore, the system incorporates a differentiable phase correlation module, which calculates the relative orientation between scan pairs through correlation analysis performed within the frequency spectrum.

Another method proposed in \cite{chen2022overlapnet}, estimates the degree of similarity between two LiDAR scans and their relative orientation. The network input is a pair of LiDAR scans, which are converted into spherical projections. The network architecture is a siamese, which has two branches with shared weights. The outputs of these branches both go to two different ones. The output is the degree of overlap between them, and the yaw angle between them.

Recently, other architectures such as MinkUNeXt \cite{cabrera2024minkunext} relies on the encoding of point clouds into global descriptors using 3D sparse convolutions, this allows the use of all the information from 3D point clouds without the need to be projected into a 2D image, so a large amount of information is captured, but in an efficient way, since sparse networks allow focusing only on the areas of the scan where there are points. Furthermore, the use of networks with U-Net architecture allows capturing the relative information between objects in the scene at different resolutions. The developed architecture has proven to be robust on large-scale datasets such as the Oxford RobotCar dataset \cite{maddern20171}, demonstrating better performance than other architectures.

\subsubsection{Fine Localization}
Fine localization can be achieved by performing a global point cloud registration. This approach provides a more accurate estimate of the position and orientation of the robot. As a consequence, a LiDAR scan can be aligned with a reference map or with a previously captured scan. The result of this process is a transformation matrix, which includes information about the 6 degrees of freedom of the robot. 

A classic method used for point cloud registration is ICP \cite{besl1992method}. This method requires an initial transformation that roughly aligns the point clouds. Then, the method performs a refinement of this transform. However, when the initial transformation differs significantly from the correct alignment between point clouds, the method does not guarantee satisfactory results. For this reason alternative global registration techniques have been developed, such as \cite{zhou2016fast}, where performance is improved without the need of a more precise initial transform.

In other papers, the alignment between point clouds is achieved by matching local features of the point clouds. These local features are obtained through the use of neural networks. In \cite{zeng20173dmatch}, a neural network capable of describing local areas is developed to set up correspondences between 3D data.

In more recent work \cite{choy2020deep}, deep neural networks are used to learn robust representations and associate corresponding points between point clouds, improving registration accuracy in the presence of misalignment and noise. These deep learning-based approaches offer a significant improvement over traditional methods, providing a more efficient and accurate solution for point cloud registration.

Other approaches \cite{ma2024ff}, take into account the registration of point clouds from different modalities, i.e., captured with different sensors. This is achieved by employing feature filtering to select points with high confidence, then detecting the local areas with the most relevant information. Finally, a transformation is obtained by optimising the alignment between the scans from local to global. This approach ensures a highly accurate registration, making it particularly useful in environments where multiple types of sensors are used.

\subsection{Localization with Monte Carlo algorithm}
The MCL algorithm is a widely used method in the field of mobile robotics to solve the kidnapped robot problem and to estimate the pose of a robot in a previously known environment. This method is based on particle filters. One of the best-recognized approaches was \cite{dellaert1999monte}, where the basis for the use of particle filters was established in the field of localization. 

The key aspect of this method is the use of a good observation model which determines, depending on the current environment captured by the robot, which weight is given to each of the particles that are distributed in a space. In some works, these observation models have relied on the appearance of images. In \cite{fernandez2012monte}, visual descriptors extracted from images are used as an observation method. However, the use of such observation methods presents challenges due to the noise in the images.

In \cite{chen2022overlapnet}, the similarity of two LiDAR scans have been used as an observation method. This method uses deep neural networks to obtain the similarity between the two point clouds. It has been shown that this type of observation method works well in large and challenging environments.

A methodology for localization in large-scale environments is proposed in \cite{sun2020localising}. The proposed method utilizes a deep neural network to learn a probability distribution of the robot's pose from bird's-eye view (BEV) images of LiDAR point clouds. This learned distribution is used to initialize the particles in MCL, instead of a random uniform distribution. This allows MCL to start with considerably more accurate pose estimation, which accelerates convergence and improves overall accuracy.

In this paper, we propose an integration of MCL with an observation model predicated on the MinkUNeXt network. MinkUNeXt's encoder-decoder architecture, coupled with its sparse nature, facilitates the extraction of detailed and differentiated information from LiDAR scans, adapting to diverse environmental configurations. Consequently, by employing a pre-existing map enriched with this information, the MCL method exhibits accelerated convergence towards the optimal solution. Furthermore, the method has significant robustness, enabling the acquisition of precise localizations even in the presence of map areas with environmental similarities but distant locations.

\section{Methodology}\label{methodology}

\begin{figure*}[!ht]
\centering
\includegraphics[scale=0.23]{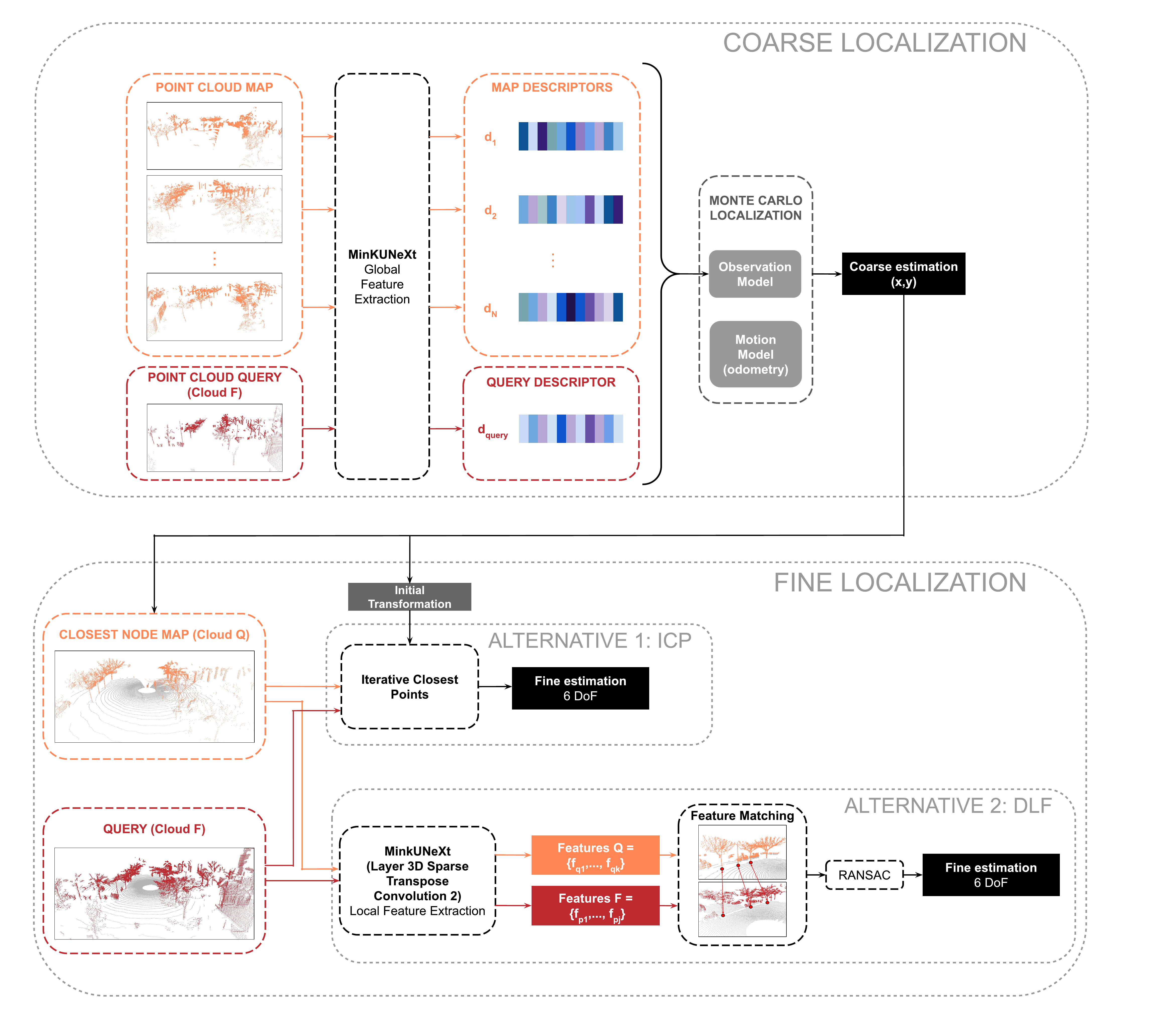}
\caption{Schematic of the proposed localization method, consisting of coarse localization and fine localization.
\textbf{Coarse localization process}: The inputs are the point clouds of the map and the query point cloud, which is derived from the current observation of the robot. From these points clouds the global descriptors are extracted with MinkUNeXt. These descriptors are used as an observation model in the MCL process. Finally, the output of this process is the estimated (x,y) position of the robot. \textbf{Fine localization process}: Two alternative methods are presented, both requiring as input the query point cloud and the closest point cloud of the map to the position (x,y) obtained with MCL. The first alternative method, is ICP, which also requires an initial transformation to relate the two point clouds in position and orientation. These data are obtained from the coarse localization process. The second alternative method, involves the use of an intermediate layer (3D Sparse Transpose Convolution 2) of the MinkUNeXt network to obtain the features of each point of the scans F and Q, which are used to do feature matching between points in both scans. The outputs of these two methods are the 6DoF estimated position of the robot.}
\label{fig:schemecoarse-fine}
\end{figure*}

The following section will provide a detailed description of the methodology employed in the localization process, including both coarse and fine localization. The coarse localization is supported by the MCL method, using as an observation model the descriptors extracted with the neural network MinkUNeXt from the 3D LiDAR scans. Once this first estimate for the robot's pose has been obtained, fine localization is carried out, for which two alternative methods are proposed. The first alternative is the classic ICP method. The integrated method resulting from the combination of MCL and ICP, is refered to throughout this paper as MCL-ICP. The second alternative is to use the output of one of the intermediate layers of the MinkUNeXt network. This approach extracts information from the scan points in order to establish the transformation between the two point clouds. The combination of the MCL method and this Deep Local Features (DLF) is denoted as MCL-DLF. The coarse and fine methods are illustrated in Figure \ref{fig:schemecoarse-fine}.

\subsection{Monte Carlo Localization method}\label{methodology_montecarlo}
\subsubsection{Description of the method}\label{methodology_montecarlo_description}

The MCL algorithm \cite{dellaert1999monte} aims to estimate the position and orientation of the robot, i.e. its state $x_t = (x, y, \theta)$ at time $t$,  utilizing data $Z_k = {z_{k,i} = 1,...,k}$ of the environment and the robot's movements $u_{1:t} = \{u_1, u_2, ..., u_t\}$. The probability density function $p(x_t|z_{1:t}, u_{1:t})$ is represented by $M$ random samples, termed particles $S_k = \{s_k^i ; i=1,...,M\}$. 

At the initialization of this method at time $t = t_0$, a random set of particles $S_0 = {s_{0}^i}$ is generated. In this paper, the MCL method has been implemented using a topological approach, so that the particles are initialized at the $(x,y)$ positions where the point clouds that constitute the map have been captured. The initial orientation of each particle is random. Consequently, each particle is given by $s_0 = \{x_0, y_0, \theta_0\}$. Once the initial set of particles is defined, the following process is performed iteratively:
\begin{itemize}
\item Prediction phase: In this phase, a new set of particles, denoted by $S_t$ and corresponding to the time $t$, is generated from the preceding set, denoted by $S_{t-1}$, and a control signal $u_t$. This new set is the result of shifting each of the particles in the set at the previous time according to the motion of the control signal. The set $S_t$ represents the density $p(x_t|z_1$:$t, u_1$:$t)$. The motion model shall be defined in Section \ref{methodology_montecarlo_motionmodel}.

\item Update phase: In this phase, the observation $z_t$ taken by the robot is used to calculate each of the weights $w_t^i$ of the particles in the set $S_t$. The process of weight assignment is described in Section \ref{methodology_montecarlo_obsmodel}. From this process, $s_t^i=(x_t^i, w_t^i)$ is obtained for each of the particles on which afterwards the resampling is performed. The resampling process is conducted randomly with the probability of selection given by the values of the particle weights. The resulting set of particles will have the same extent $M$ as the input set, and will be composed of the particles with the highest weights.
\end{itemize}

Once these two procedures have been carried out, the estimated position of the MCL method will be given by Equation \ref{e:part_media} as the mean position of the set of particles of that iteration. 

\begin{equation} \label{e:part_media}
\hat{s}_{t}=\frac{1}{M} \sum_{i=1}^{M} s^{[i]}_t
\end{equation}

Subsequently, the process continues iteratively. In order to obtain a more accurate position, the position $\hat{s}_{t}$ obtained in each iteration is refined in accordance with the process described in Section \ref{methodology_fineloc}.

\subsubsection{Motion Model}\label{methodology_montecarlo_motionmodel}
In the prediction phase of the MCL method, the particles are propagated with the odometry data. The relative odometry between the previous and the current pose is thus considered. With these relative odometry defined as $u_t= (\Delta x_t, \Delta y_t, \Delta\theta_t)$. The distance travelled by the robot on the basis of this odometry will be $d_t=\sqrt{\Delta x_t²+\Delta y_t²}$. Therefore, the new state of the particle is obtained according to Equations \ref{e:x_model}, \ref{e:y_model} y \ref{e:theta_model}.
\vspace{-0.3cm}
\begin{equation} \label{e:x_model}
x_{t}= x_0 + d_t * cos( \theta_0 +  \Delta\theta_t)
\end{equation}
\vspace{-0.8cm}
\begin{equation} \label{e:y_model}
y_{t}= y_0 + d_t * sin( \theta_0 +   \Delta\theta_t)
\end{equation}
\vspace{-0.8cm}
\begin{equation} \label{e:theta_model}
\theta_{t}= \theta_0 + \Delta\theta_t
\end{equation}
Once these data have been obtained and following the perspective of the topological method. Each particle will be approximated in $x$ and $y$ with the position of the nearest node. 

\subsubsection{Observation model}\label{methodology_montecarlo_obsmodel}

The MCL method evaluates the probability of each generated particle in the set being at the current position of the robot. Therefore, this measure reflects the proximity of each particle to the actual pose of the robot. To conduct this evaluation, MCL employs an observation model. This observation model assigns a weight to each particle, based on a comparison of the features of the environment observed by the robot and those of the map. In this case, the study of the features of the environment is carried out on the basis of the point clouds obtained with a LiDAR sensor.  

The point clouds have a great amount of information. Consequently, the process of comparing scans to assess their similarity would necessitate a significant computational cost. Therefore, we have opted to utilize deep learning techniques that facilitate the embedding of the point clouds, thereby ensuring the preservation of the most salient information and significantly reducing the time required for computational comparisons. In this paper, the MinkUNeXt \cite{cabrera2024minkunext} architecture is employed, a neural network capable of efficiently encoding extensive three-dimensional geometric environments.

The main objective of the MinkUNeXt architecture is to solve the problem of place recognition using point clouds, for which this neural network uses 3D sparse convolutions. Feature extraction is performed at different scales according to a U-net architecture with an encoder-decoder topology. In the encoder section, a progressive reduction of resolution is done to obtain from finer to more general features. In the decoder section, a progressive upsampling is implemented, reconstructing the point cloud until the original resolution. Also, in the decoder, skip connections are performed with the encoder to merge general and fine features. After the decoder, there is a fully connected layer to provide the descriptors of each point in the cloud with invariance to viewpoint changes. Finally, the feature map passes through a Generalized Mean Pooling layer, where it is embedded in a single descriptor of length 512. In this way, thanks to the MinkUNeXt architecture, it is possible to describe the relevant information of a point cloud in a single vector. The layers of this network employ 3D sparse convolutions, enabling the efficient capture of relevant information from the point cloud. This network is implemented using the Minkowski Engine library. \cite{choy20194d}

This architecture is employed to calculate the map descriptors, denoted as $D_{map}=\{d_1,..,d_N\}$ and the robot descriptor $d_{query}$. $d_{query}$  is derived from the point cloud captured by the robot at time $t$. Conversely, each descriptor in the set $D_{map}$ is generated from a point cloud of the map. The map consists of $N$ two-dimensional nodes $\{(n_{1,x}, n_{1,y}),...,(n_{N,x}, n_{N,y})\}$, each containing a point cloud that has been captured at the position where the node is located.

During the robot's localization process, the current robot descriptor $d_{query}$, is compared against all the map descriptors in $D_{map}$ to assess the similarity between the robot's current perception and the map's representations. Subsequently, the $B$ map descriptors that exhibit the highest similarity to $d_{query}$, as determined by the shortest Euclidean distance in the descriptor space, are selected.

The weight of each particle is determined by two distances: the metric distance, $v_{j}$,  and the distance in the descriptor space $h_{j}$. 

The metric distance is calculated as $v_{j}=(n_{k,x}, n_{k,y}) - (x_i, y_i)$, where $(n_{k,x}, n_{k,y})$ is the position of one of the B nodes of the map and $(x_i, y_i)$ is the position of the particle i. 

Alternatively, the distance of the descriptors can be expressed as $h_{j}=|d_k - d_i|$, where $d_k$ is the descriptor of one of the $B$ nodes of the map and $d_i$ is the descriptor of the closest node of the particle in the map.

\begin{figure*}[ht]
\centering
    \begin{subfigure}[b]{0.35\textwidth}
        \centering
        \includegraphics[width=\textwidth]{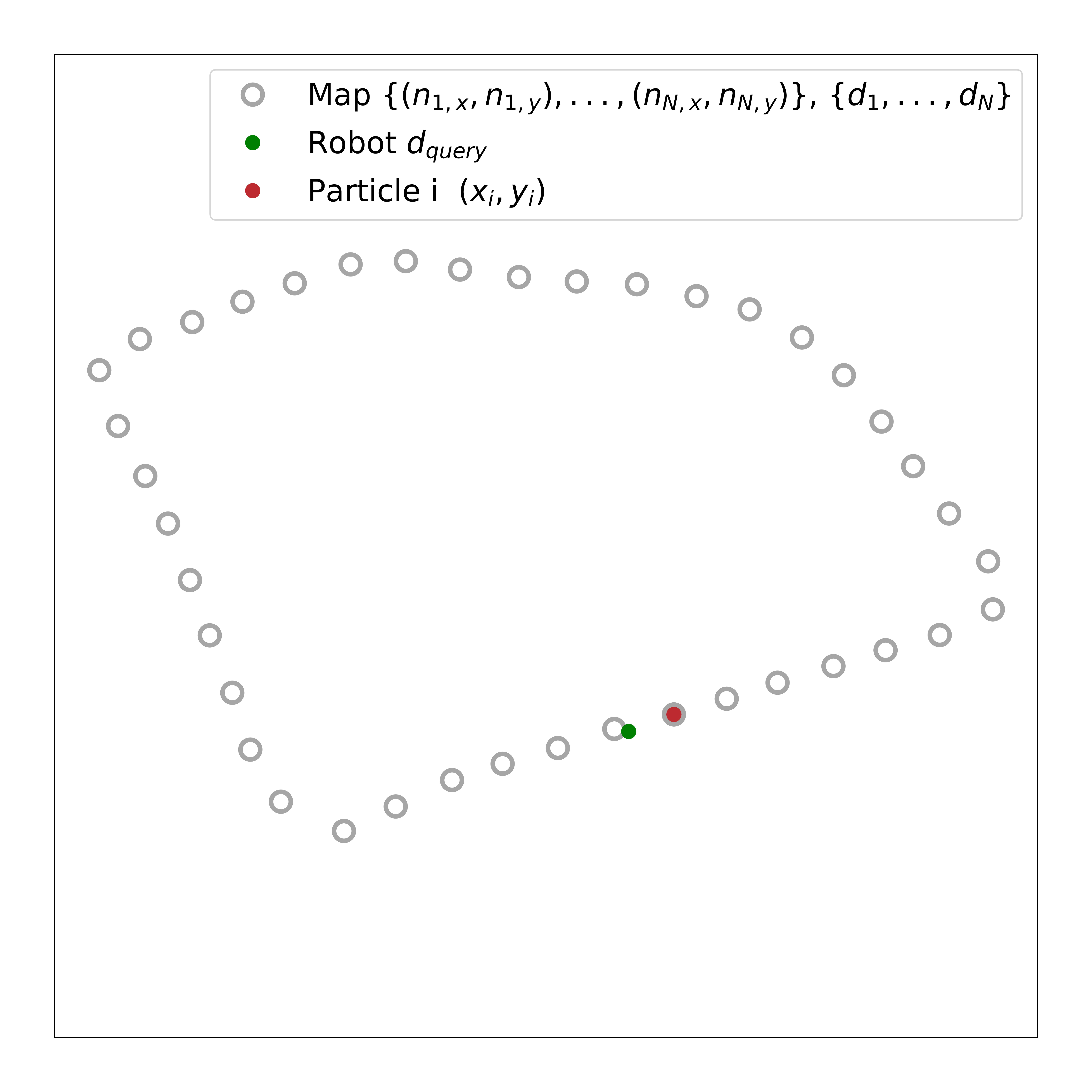}
        \caption{Initial conditions}
        \label{fig:scheme_weights1}
    \end{subfigure}
    \begin{subfigure}[b]{0.35\textwidth}
        \centering
        \includegraphics[width=\textwidth]{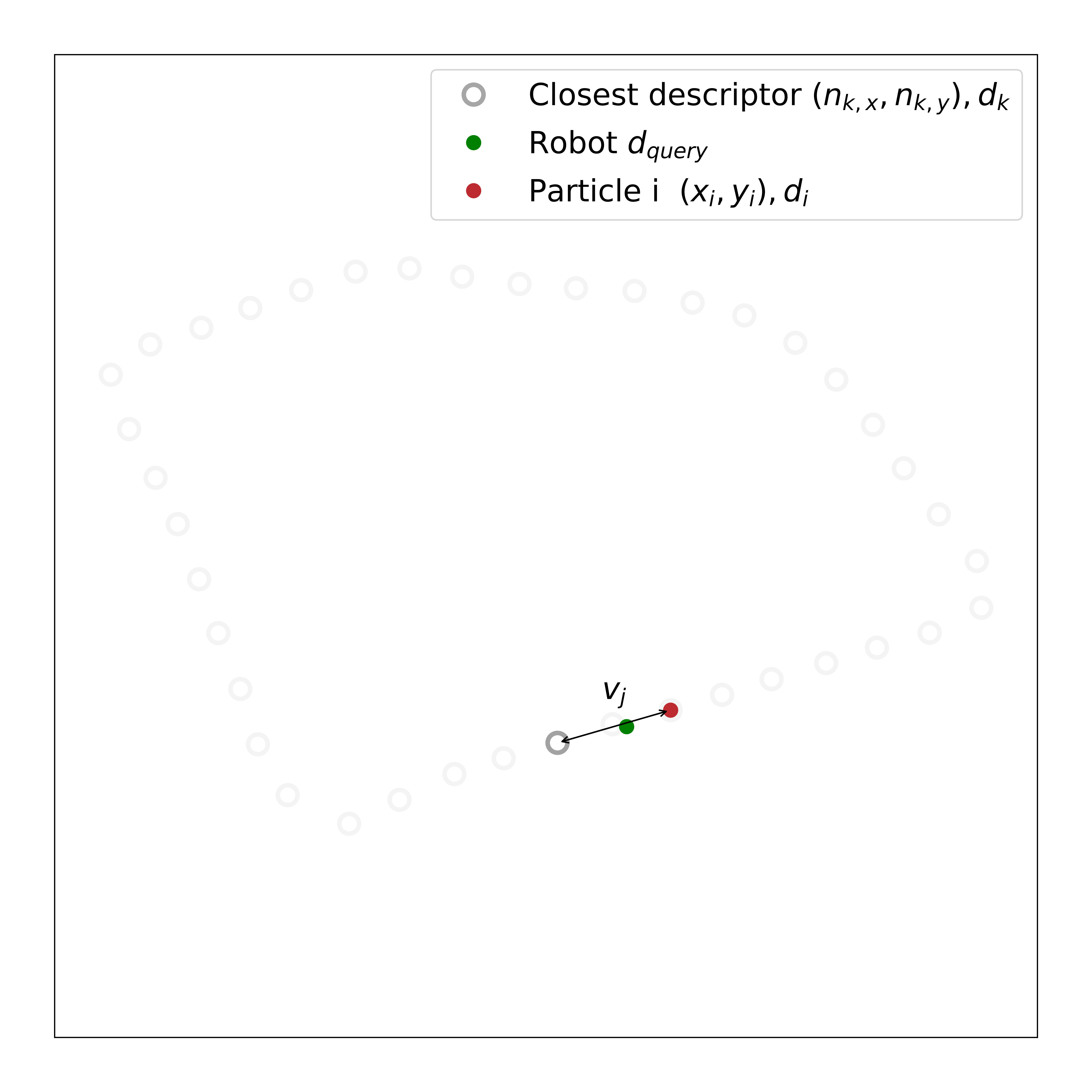}
        \caption{Weight process}
        \label{fig:scheme_weights2}
    \end{subfigure}
\caption{Weighting process of particle i. \textbf{(a)} the map nodes, located at positions $\{(n_{1,x}, n_{1,y}),...,(n_{N,x}, n_{N,y})\}$, are visualized as grey circles. Particle i, at coordinates $(x_i, y_i)$, is shown in red, and the robot, for which we only know the descriptor $d_{query}$ from the LiDAR observation, is shown in green. This descriptor is used to compare with all map descriptors $\{d_1,..,d_N\}$. \textbf{(b)} the map node with the closest descriptor is shown as a grey circle. Then, $v_j$ is calculated as the difference between the node's position $(n_{j,x}, n_{j,y})$ and the particle's position $(x_i, y_i)$. Finally, $h_j$ is calculated as the difference between $d_k$ and $d_i$, where $d_i$ is the descriptor of the map node where the particle $i$ is located.}
\label{fig:weights_particle_i}
\end{figure*}

The matrices $\Sigma_l = diag(\sigma_l^2, \sigma_l^2)$  and $\Sigma_m =1/\sigma_m$  are also employed to model the adjustment of the weights. The total weight of each particle is calculated in Equation \ref{e:particles_w} and the described process can be observed in Figure \ref{fig:weights_particle_i}.

\begin{equation} \label{e:particles_w}
w_{t}^i= \sum_{j=1}^{B}{exp(-v_j\Sigma_l^{-1}v_j^T)exp(-h_j\Sigma_m^{-1}h_j^T)} 
\end{equation}

\begin{figure*}[h]
    \centering
    \begin{subfigure}[b]{0.24\textwidth}
        \includegraphics[width=\textwidth]{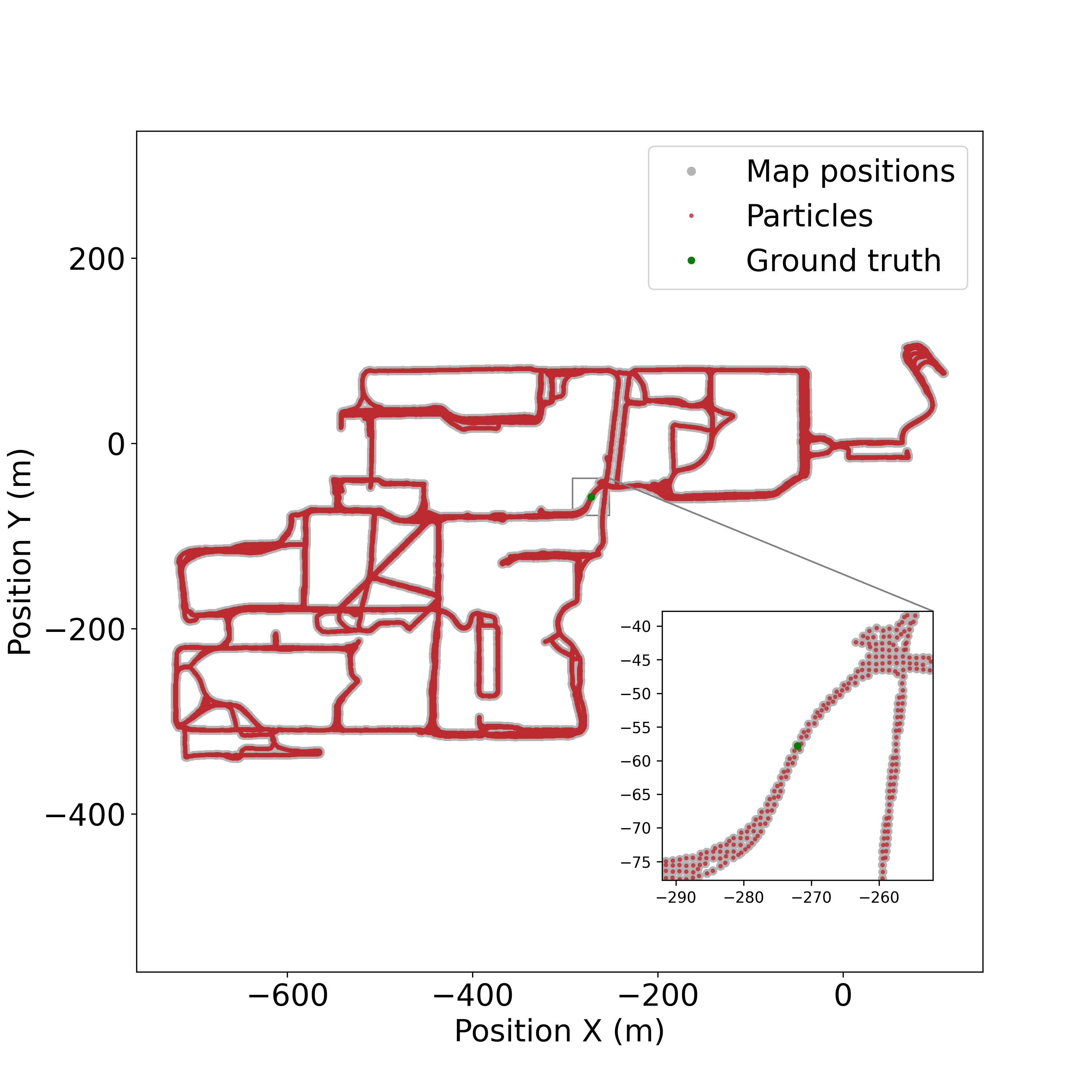}
        \caption{Iteration 0}
        \label{fig:dispersion1}
    \end{subfigure}
    \begin{subfigure}[b]{0.24\textwidth}
        \includegraphics[width=\textwidth]{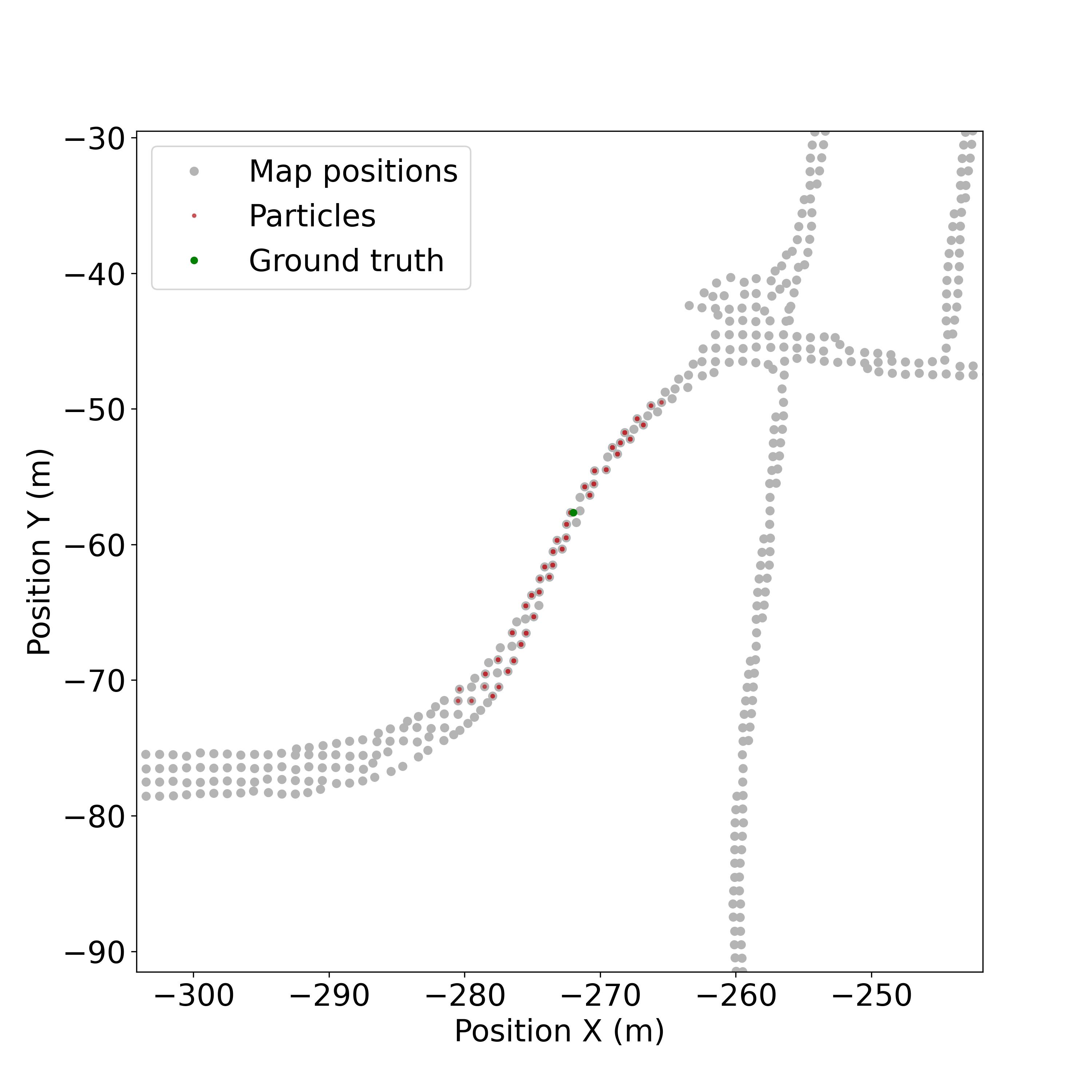}
        \caption{Iteration 1}
        \label{fig:dispersion2}
    \end{subfigure}
    \begin{subfigure}[b]{0.24\textwidth}
        \includegraphics[width=\textwidth]{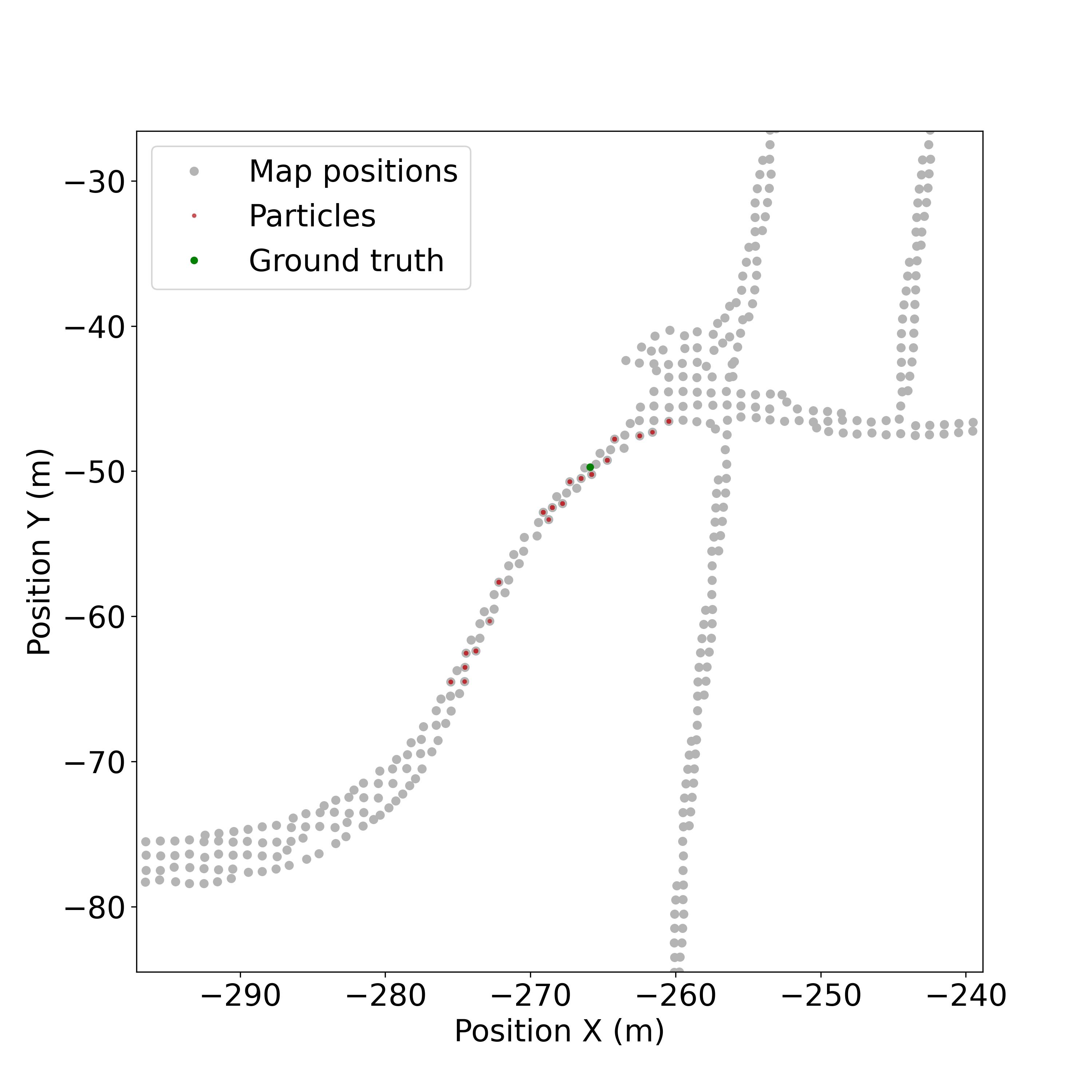}
        \caption{Iteration 10}
        \label{fig:dispersion3}
    \end{subfigure} 
    \begin{subfigure}[b]{0.24\textwidth}
        \includegraphics[width=\textwidth]{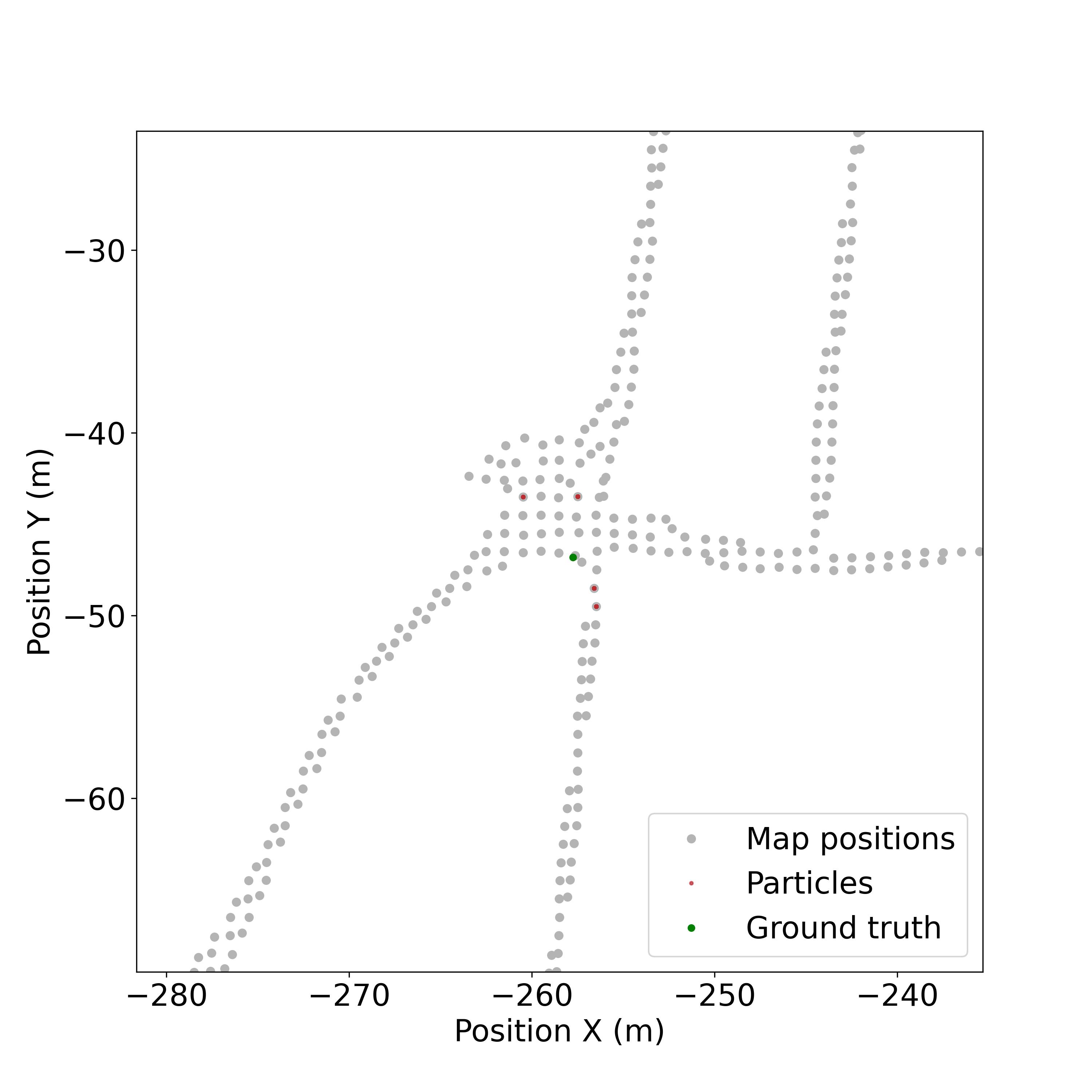}
        \caption{Iteration 20}
        \label{fig:dispersion4}
    \end{subfigure}
    \caption{Particle locations, in red, relative to the map positions, in grey, and the robot's current location, in green. \textbf{(a)} the particles are shown distributed across all map nodes. \textbf{(b)} and \textbf{(c)} the particles converge to a smaller area of the map. \textbf{(d)} the particles have converged to very close areas near the robot's actual location.}
    \label{fig:dispersionofparticles}
\end{figure*}

Thus, the observation process allows to weight each particle in the set and perform successive iterations as the robot moves. The convergence process of the particles during MCL iterations is illustrated in Figure \ref{fig:dispersionofparticles}.

\subsection{Fine localization}\label{methodology_fineloc}

Once the global localization has been carried out using the MCL method with the proposed observation model, we proceed to estimate the pose more precisely. 

The initial location obtained through the coarse method will be utilized for the subsequent procedure. The point cloud of the map that is closest to this initial location will be employed as the reference point for the subsequent analysis. With this point cloud and the one captured by the robot at the current time, pairwise registration will be performed. In this paper, two alternative methods have been evaluated: ICP and deep local features from the intermediate layers of the MinkUNeXt network.

\subsubsection{Iterative Closest Point (ICP)}

In this case, the coarse localization phase is first addressed through the MCL process described in the preceding section, and subsequently, the ICP method is employed for fine localization; this combined global method is termed MCL-ICP (Monte Carlo Localization - Iterative Closest Point).

ICP is a classic registration algorithm. To start with, it needs a homogeneous transformation matrix T that roughly relates the two point clouds: target ($Q$) and source ($F$). This method allows the matrix to be refined to improve the alignment between these two point clouds.

The ICP method consists in finding correspondences $K=\{(p,q)\}$, where $p$ and $q$ are points of the point clouds $P$ and $Q$ respectively. 

Based on these correspondences, we then update the transformation $T$ by minimizing a cost function $E(T)$, which is specific to the chosen variant of the ICP algorithm and its implementation. In this case, Point-to-plane ICP has been used, where $E$ is defined according to Equation \ref{e:icp_error} (where $n_p$ is the normal of point $p$).

\begin{equation} \label{e:icp_error}
E(T)= \sum_{(p,q)\epsilon K}{((p-T_q) n_p)²} 
\end{equation}

\subsubsection{Deep Local Features (MinkUNeXt)}

In this method, coarse localization is also achieved using MCL, and subsequently, fine localization is performed utilizing deep local features; this complete method is termed MCL-DLF (Monte Carlo Localization - Deep Local Features).

This fine localization is achieved by extracting the information from the intermediate layers of the MinkUNeXt network. The study of the output provided by a sparse encoder-decoder network, such as MinkUNeXt, for representing 3D geometric spaces has been conducted, achieving high accuracy through the use of specific intermediate layers, without the necessity of training for this specific task.

The information from these intermediate layers shall consist of descriptors for each point in the point cloud. The point cloud's size and the length of its descriptors rely on the output of the layer being used.

Once we have obtained the descriptors of the two point clouds, we will proceed to calculate the correspondences between the points of both point clouds. Considering the feature points of two scans, $F$ and $Q$, represented by the sets $F_F =\{f_{p1}, f_{p2},..., f_{pj}\}$ (where $j$ denotes the total number of points in $F$) and $F_Q =\{f_{q1}, f_{q2},...,f_{qk}\}$ (where $k$ denotes the total number of points in $Q$), we generate correspondences $K=\{(p,q)\}$ based on their Euclidean distances. These correspondences, along with the original point clouds, are then used by the RANSAC \cite{fischler1981random} method to estimate the transformation between the scans.

\section{Experiments}
\subsection{Datasets}\label{section:experiments_datasets}
In order to evaluate the proposed method, data from the University of Michigan North Campus Long-Term Vision and LiDAR Dataset (NCLT) \cite{carlevaris2016university} dataset have been utilized. Furthermore, experiments have been carried using own data captured at the Miguel Hernández University of Elche. These datasets have been selected due to they were captured in challenging environments. They encompass large-scale settings exhibiting significant variability, particularly seasonal changes. Furthermore, they include handover scenarios involving transitions between outdoor and indoor environments.

\subsubsection{NCLT}
The NCLT dataset \cite{carlevaris2016university} consists of data from different sensors integrated in the Segway robotic platform. These sensors include: omnidirectional imagery, 3D LiDAR, planar LiDAR, GPS, proprioceptive sensors and odometry. Specifically, the 3D LiDAR sensor is the Velodyne HDL-32E LiDAR. 
 
The odometry is estimated with an Extended Kalman Filter (EKF) that fuses data from the robot's wheel encoder, a single-axis Fibre Optic Gyro (FOG) and an Inertial Measurement Unit (IMU). In addition, this dataset also provides ground-truth pose data for all sessions generated via SLAM.

The data have been captured in indoor and outdoor areas of the University of Michigan's North Campus. It contains 27 mapping sessions, captured on different days and at different times over a period of 15 months.

This variety of routes has enabled the acquisition of scenarios with major changes, such as alterations in lighting, variations in tree foliage, and the presence or absence of snow on the streets. Additionally, there are other elements that are subject to change, including obstacles such as people or bicycles. 

\subsubsection{University Miguel Hernandez of Elche (UMH)}

The second dataset consists of information related to routes recorded on the Miguel Hernandez University of Elche campus \footnote{Data is available at \url{https://arvc.umh.es/db/databases/}}. These routes have been acquired with a Husky A200 robot. The main purpose of recording these routes is to test navigation, mapping and localization algorithms for mobile robots. The true positions of the routes have been obtained, these will be utilized to calculate the error when compared to the poses obtained through our localization method. The true positions have been obtained through the use of a SLAM process. The acquisition of point clouds has been enabled by the integration of the Ouster OS1-128 LiDAR within the robotic platform.

\begin{figure*}[t]
    \centering
    \begin{subfigure}[b]{0.30\textwidth}
        \includegraphics[width=\textwidth]{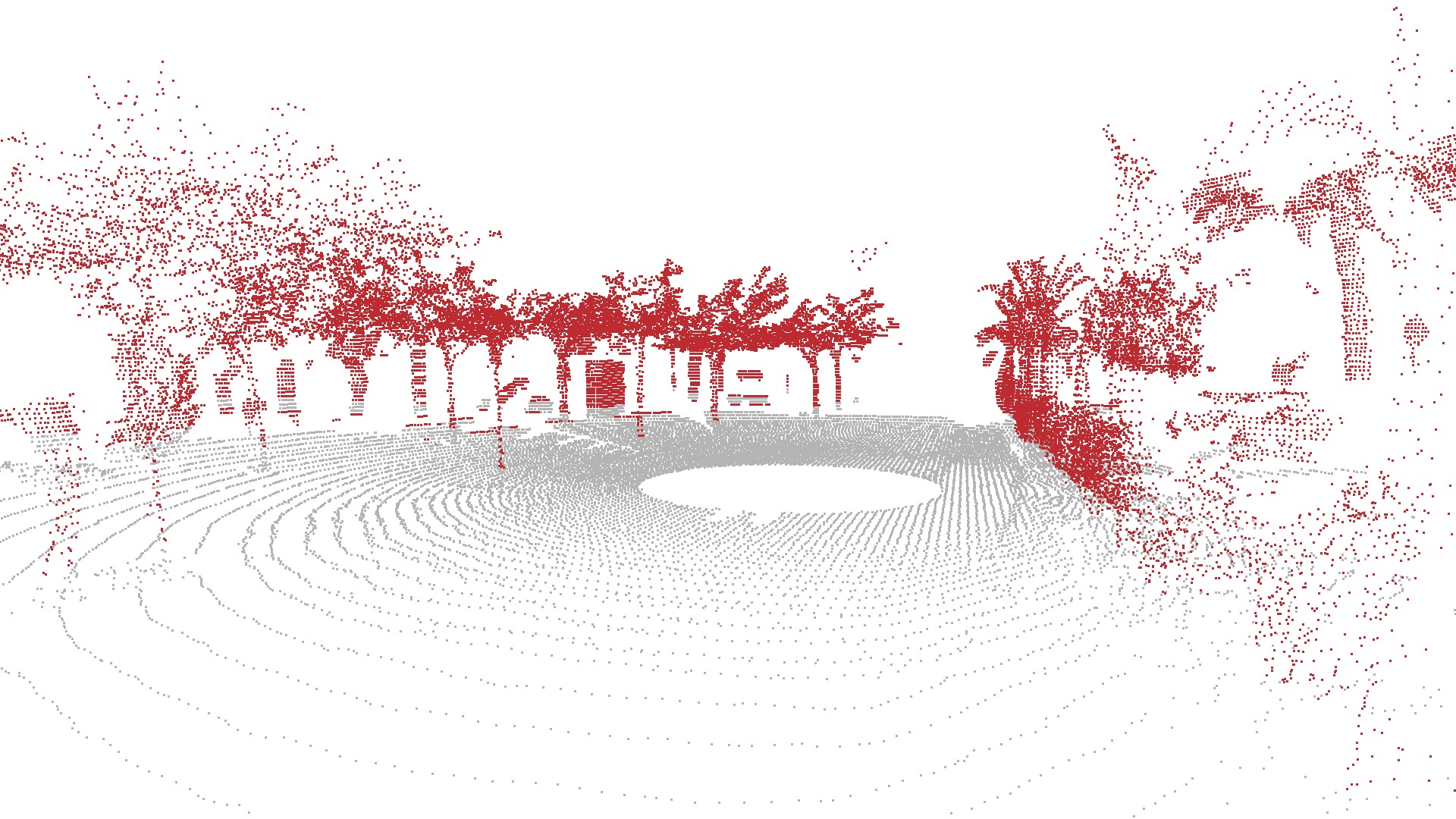}
        \caption{Outdoor point cloud}
        \label{fig:images_environment_umh_sub1}
    \end{subfigure}
    \begin{subfigure}[b]{0.275\textwidth}
        \includegraphics[width=\textwidth]{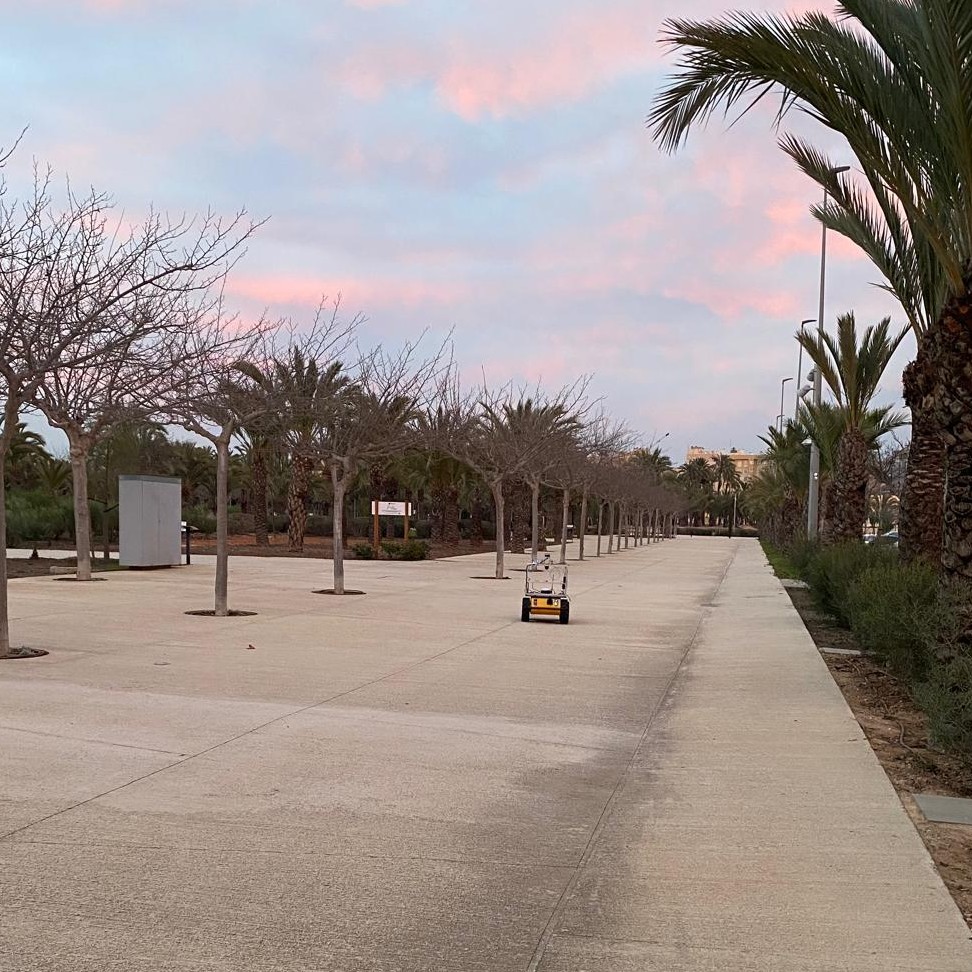}
        \caption{Outdoor environment}
        \label{fig:images_environment_umh_sub2}
    \end{subfigure}
    \begin{subfigure}[b]{0.279\textwidth}
        \includegraphics[width=\textwidth]{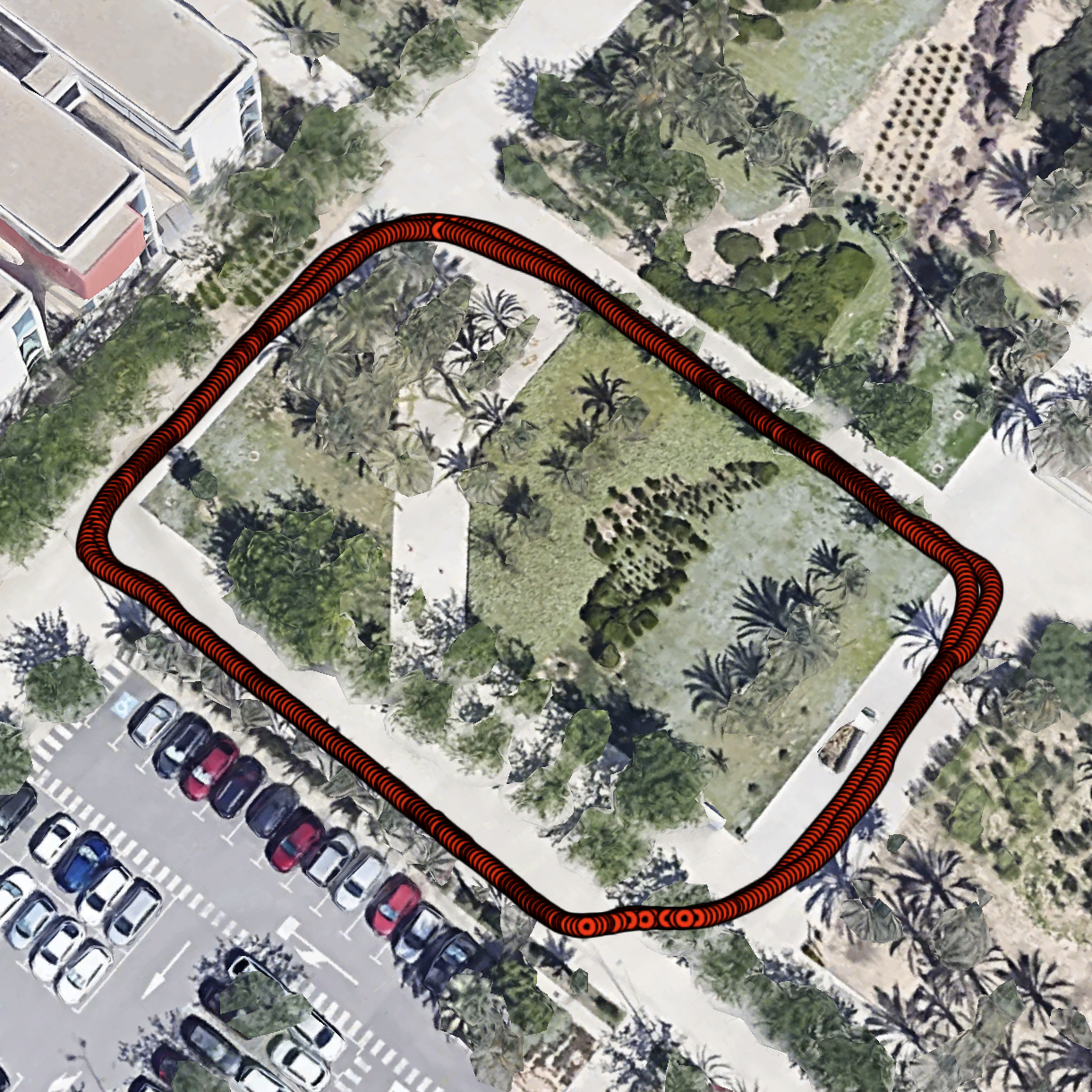}
        \caption{Satellite view of outdoor environment}
        \label{fig:images_environment_umh_sub3}
    \end{subfigure} 
    \begin{subfigure}[b]{0.30\textwidth}
        \includegraphics[width=\textwidth]{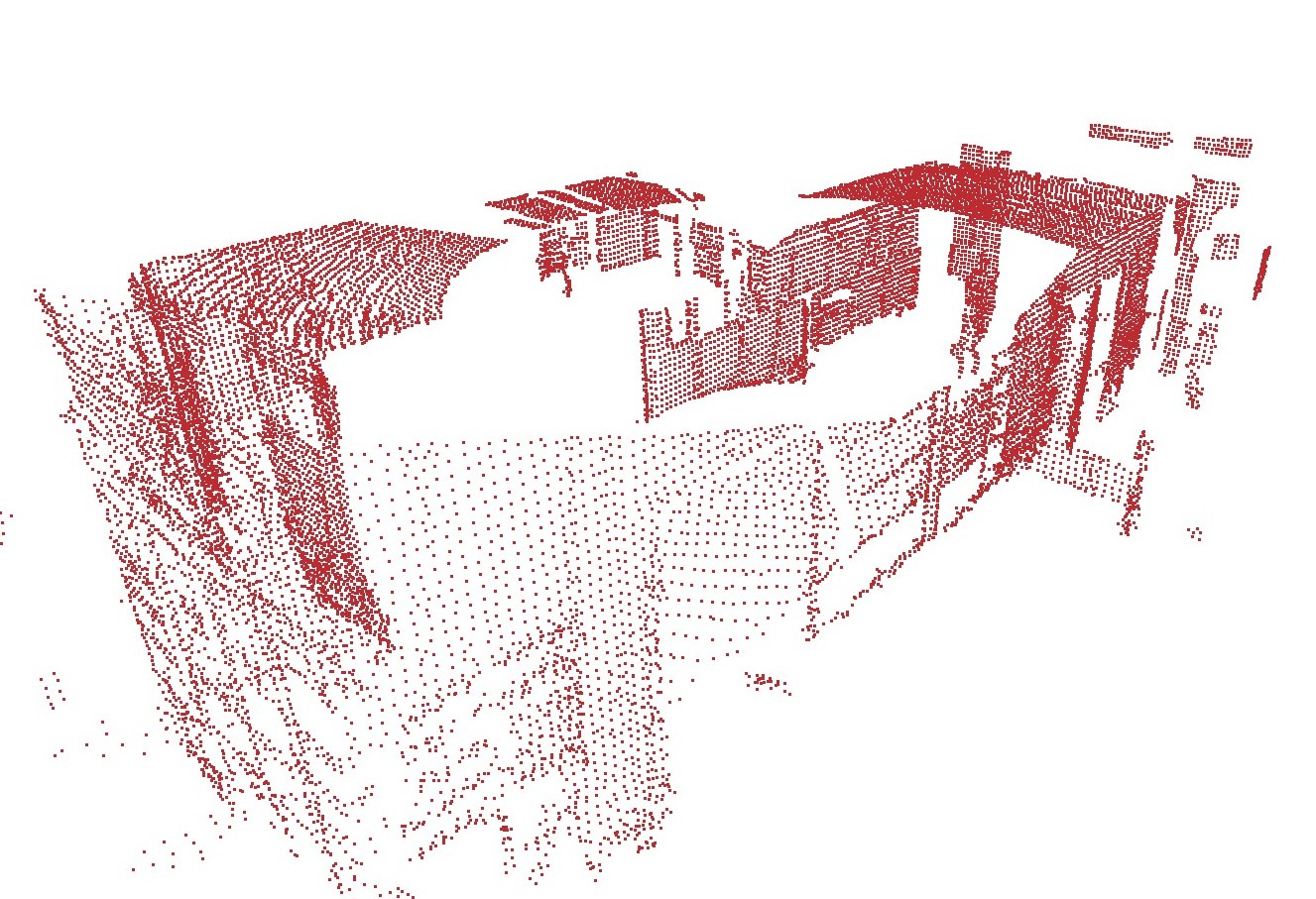}
        \caption{Indoor point cloud}
        \label{fig:images_environment_umh_sub4}
    \end{subfigure}
    \begin{subfigure}[b]{0.266\textwidth}
        \includegraphics[width=\textwidth]{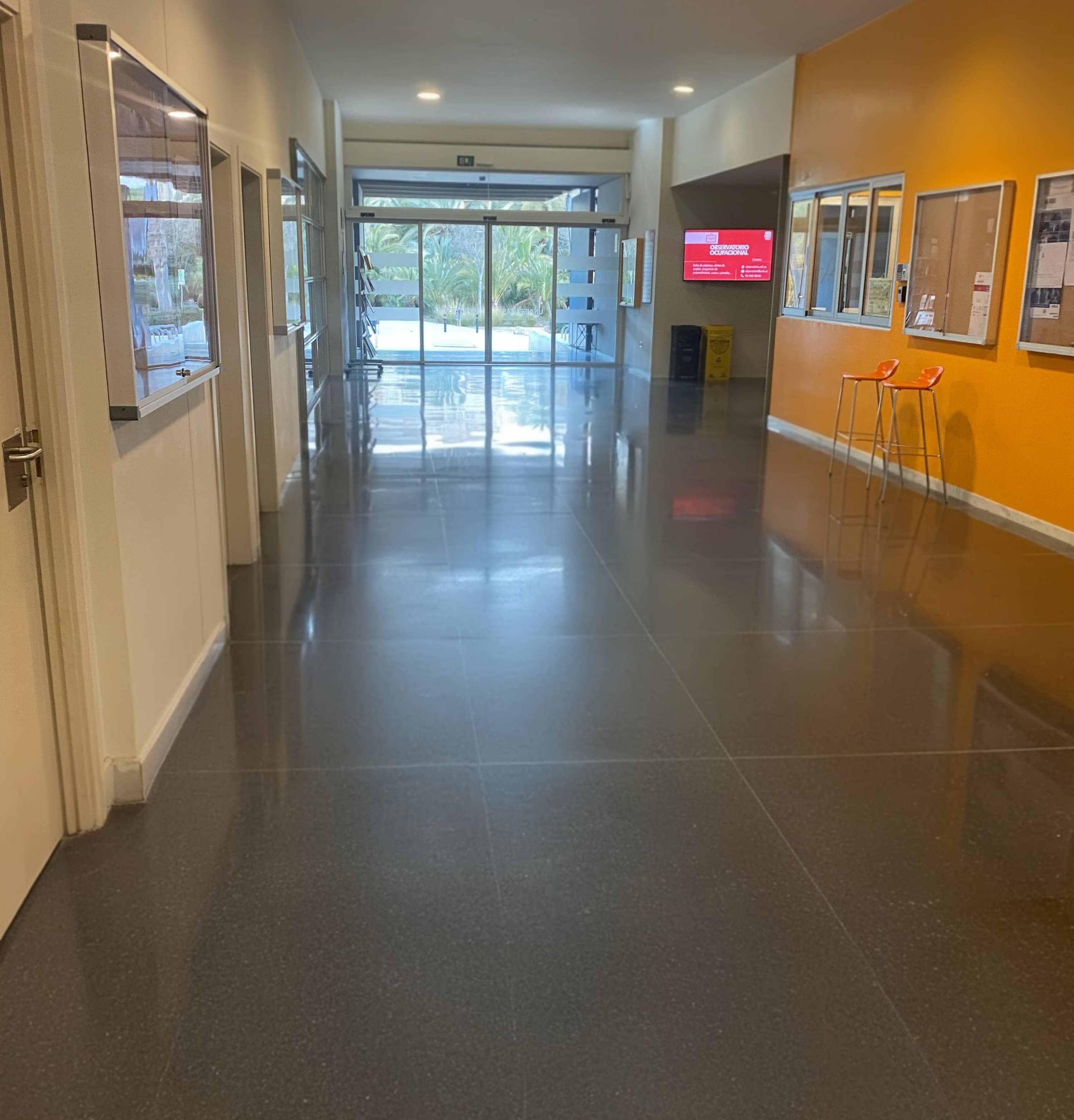}
        \caption{Indoor environment}
        \label{fig:images_environment_umh_sub5}
    \end{subfigure}
    \begin{subfigure}[b]{0.277\textwidth}
        \includegraphics[width=\textwidth]{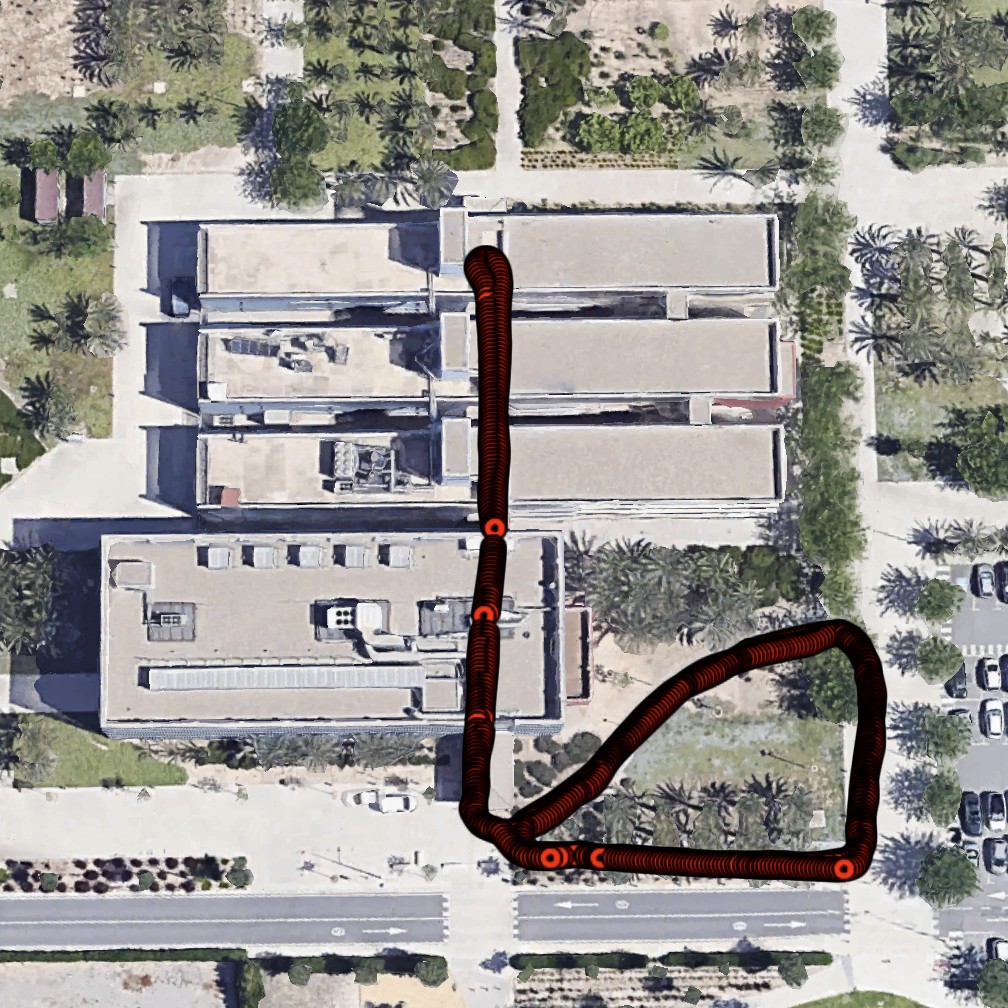}
        \caption{Satellite view of indoor environment}
        \label{fig:images_environment_umh_sub6}
    \end{subfigure}
    \caption{Appearance of the dataset environments captured at UMH. \textbf{(a) (b) (c)} outdoor environments and \textbf{(d) (e) (f)} indoor environments. \textbf{(c)} a trajectory captured entirely in an outdoor environment is shown. \textbf{(f)} a trajectory captured partly outdoors and partly indoors is presented.}
    \label{fig:images_environment_umh}
\end{figure*}

The routes in this environment comprise both indoor and outdoor areas as illustrated in Figure \ref{fig:images_environment_umh}. The outdoor areas refer to the green spaces and structures present within the campus environment. Meanwhile, the indoor areas correspond to the indoor spaces within university buildings.

\begin{figure*}[h!]
\centering
\includegraphics[width=0.75\textwidth]{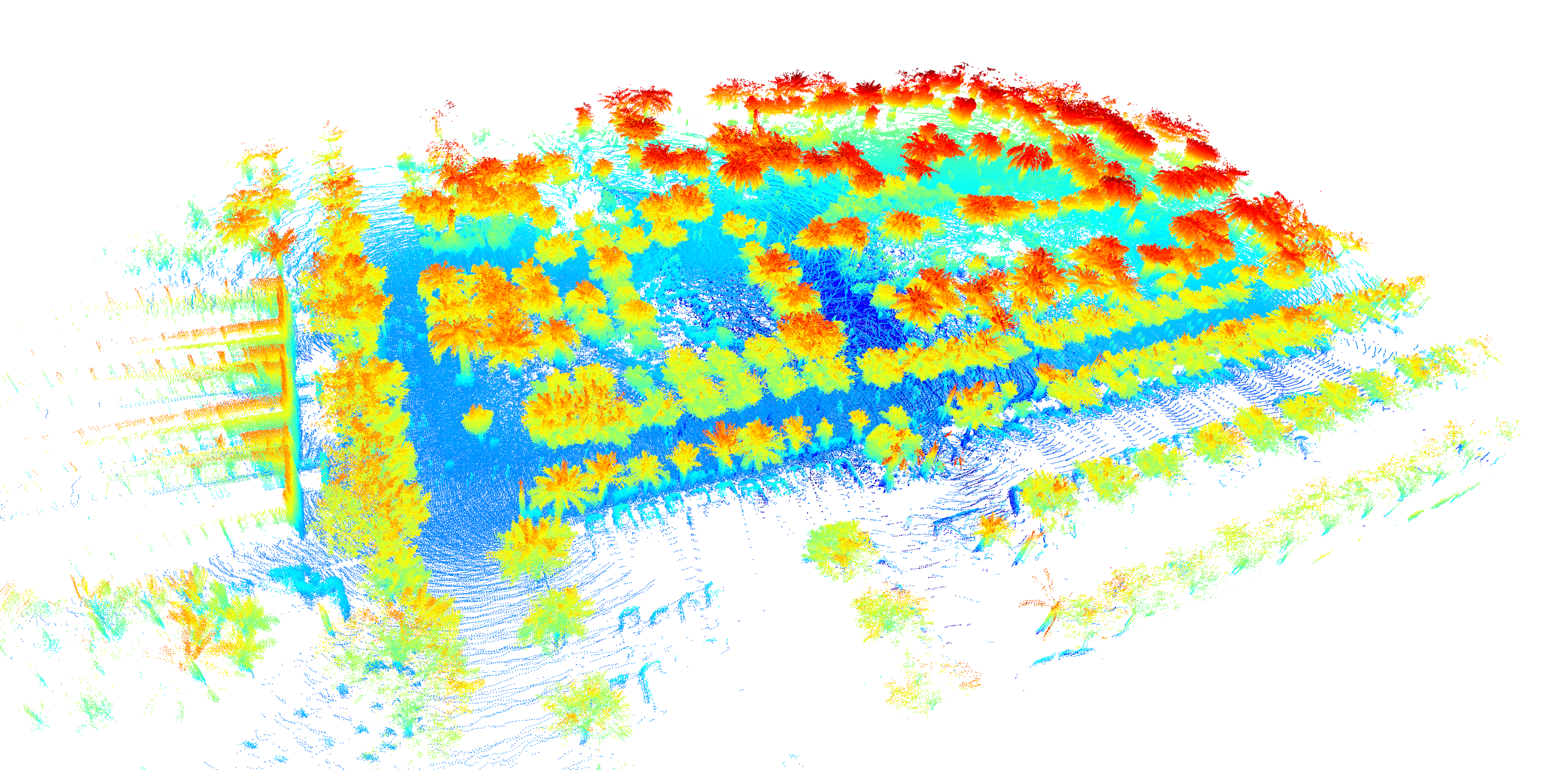}
\caption{Aggregated point clouds of a robot trajectory in the campus of Miguel Hernandez University of Elche.}
\label{fig:pcd_map}
\end{figure*}

These data were collected over several months in different seasons, allowing the observation of changes in the environment, including variations in the foliage of the trees. Additionally, other dynamic changes, such as the presence of people, bicycles or cars can be discerned. The point clouds of one captured session are illustrated in Figure \ref{fig:pcd_map}.

\subsection{Implementation Details}
In MinkUNeXt \cite{cabrera2024minkunext}, the training process was performed using the Oxford RobotCar dataset and a dataset proposed in \cite{uy2018pointnetvlad}. The data from these databases, used for training in \cite{cabrera2024minkunext}, contain scans with only 4096 points, unlike the NCLT and UMH databases, which have higher point count. To generalize to these types of scans, the weights obtained from \cite{cabrera2024minkunext} were used to perform transfer learning with the NCLT and UMH databases. The trajectories that have been selected for training the network are shown in Tables \ref{t:nclt_sessions} and \ref{t:umh_sessions}.

The clouds obtained with the LiDAR sensors have been processed to make the inferences for description extraction. During this preprocessing stage, points located at a distance exceeding 50 meters from the LiDAR origin are removed. The justification for this is that beyond that range, the point density becomes insufficient due to the LiDAR's operational limitations. After this, a normalization step is performed to center the point cloud at the LiDAR sensor's origin. Subsequently, the points are spatially scaled using an optimal scale factor of 50. This scaling ensures that all points are mapped to the range [-1, 1], given that no points are located beyond 50 meters from the origin. In addition, a downsampling process is implemented, in which the points within the point cloud are significantly reduced. In the next step, the points belonging to the ground plane are removed, as they do not provide valuable information.

A similar procedure is employed to process the point clouds to perform the point cloud registration method. The normalization step is not carried out in this case, because this would modify the dimensionality of the point cloud.

\begin{figure*}[h]
    \centering
    \begin{subfigure}[b]{0.60\textwidth}
    \includegraphics[height=6cm]{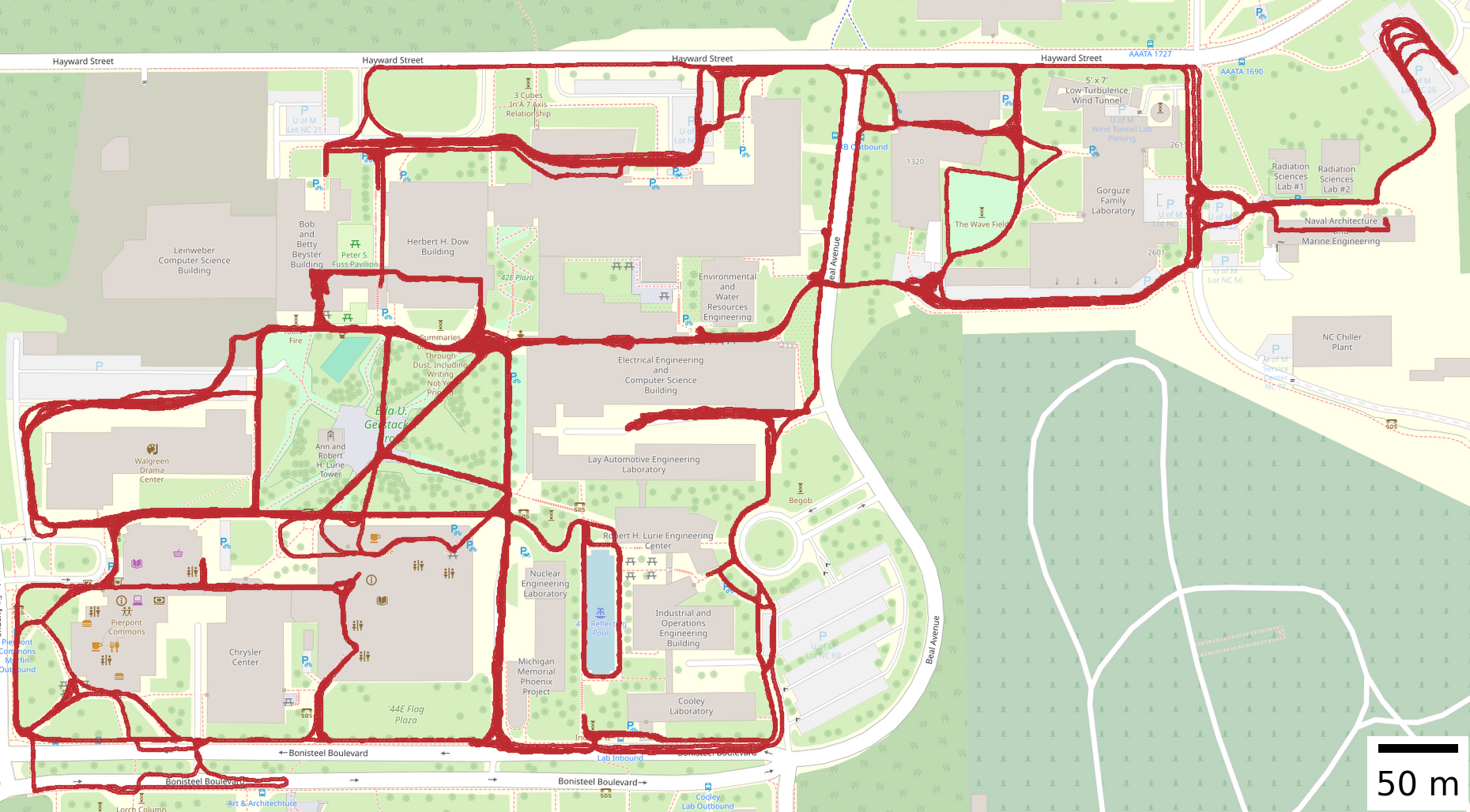}
        \caption{NCLT}
        \label{fig:maps_osm_sub1}
    \end{subfigure}
    \begin{subfigure}[b]{0.30\textwidth}
    \centering
    \includegraphics[height=6cm]{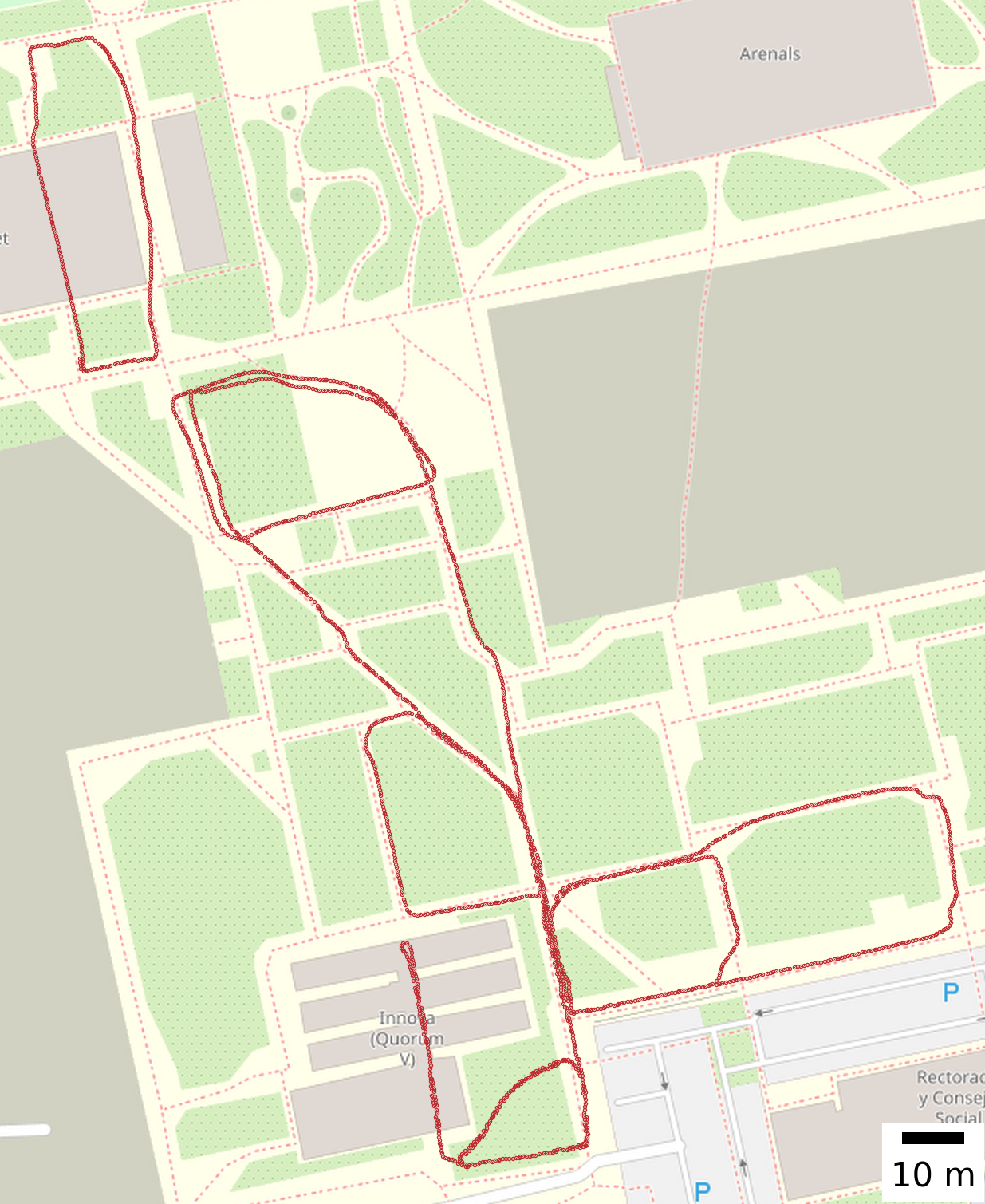}
        \centering \caption{UMH}
        \label{fig:maps_osm_sub2}
    \end{subfigure}
    \caption{Satellite view in OpenStreetMap of the data included in the maps of the \textbf{(a)} NCLT and \textbf{(b)} UMH datasets.}
    \label{fig:maps_osm}
\end{figure*}

A reference map has been used in the experiments. In each case, the formation of the map has been determined by the same trajectories employed during the retraining of the network. Therefore, for each dataset, the map is given by the trajectories indicated in Tables \ref{t:nclt_sessions} and \ref{t:umh_sessions}. The formation of the map comprises the point clouds and the positions of each of them. Another fact is that the nodes of these maps are spaced every 1 meter, thus reducing the computational expense in the MCL method. Figure \ref{fig:maps_osm} illustrates these maps, showing the points that form the map from a satellite view.

\newpage

In order to evaluate the proposed method, experiments have been carried out with different routes. These differ from the routes that constitute the map. These routes are presented in detail in Tables \ref{t:nclt_sessions} and \ref{t:umh_sessions}.

The following section details the results of experiments conducted using the proposed method. In these experiments, the MCL method is employed so that the particle update is performed each time the robot has travelled one meter. The method is initialized after 20 iterations. The position is considered to be non-knowledgeable, so the particles are re-initialized at all map nodes.

\begin{center} 
\begin{table}[H]
\caption{Data sessions from University of Michigan North Campus Long-Term vision and LiDAR Dataset used for network training, mapping and experiments.\label{t:nclt_sessions}}
\begin{tabular*}{\columnwidth}{@{\extracolsep\fill}ccc@{}}
    \toprule
    \textbf{Date [YYYY-MM-DD]} & \textbf{Length [km]}  & \textbf{Usage}\\
    \midrule
    2012-01-08 & 6.4 & Train and map\\
    2012-01-15 & 7.5 & Train and map\\
    2012-01-22 & 6.1 & Train and map\\
    2012-02-02 & 6.2 & Train and map\\
    2012-02-04 & 5.5 & Train and map\\
    2012-03-31 & 6.0 & Train and map\\
    2012-02-18 & 6.2 & Experiment February\\
    2012-04-29 &  3.1 & Experiment April \\
    2012-05-11 & 6.0 & Experiment May \\
    2012-06-15 & 4.1 & Experiment June  \\
    2012-08-04 & 5.5 & Experiment August  \\
    2012-10-28 & 5.6 & Experiment October \\
    2012-11-04 & 4.8 & Experiment November \\
    2012-12-01 & 5.0 & Experiment December \\
    \bottomrule
\end{tabular*}
\end{table}
\end{center}
\vspace{-1cm}
\begin{center} 
\begin{table}[H]
\caption{Data Sessions from University of Miguel Hernandez University data used for network training, mapping and experiments.\label{t:umh_sessions}}
\begin{tabular*}{\columnwidth}{@{\extracolsep\fill}ccc@{}}
    \toprule
    \textbf{Date [YYYY-MM-DD]} & \textbf{Length [m]}   & \textbf{Usage}\\
    \midrule
    2024-04-24 & 481.40 & Train and map\\
    2024-05-03 & 472.30 & Train and map\\
    2024-05-07 & 299.70 & Train and map\\
    2024-05-14 & 473.00 & Train and map\\
    2024-04-24 & 346.40 & Experiment April\\
    2024-06-20 & 303.68 & Experiment June \\
    2025-01-21 & 171.19 & Experiment January\\
    \bottomrule
\end{tabular*}
\end{table}
\end{center}

Furthermore, regarding fine localization estimate, ICP and registration with deep local features is tested in these experiments. On the other hand, deep local features method does not require a prior initial transformation, whereas ICP does. In order to form the transformation, it is necessary to have $x$, $y$, $z$ position, as well as $roll$, $pitch$ and $yaw$ orientation. We assume that the robot is above and parallel to the ground, so we consider $z$, $roll$ and $pitch$ as zero. The values of $x$ and $y$ will be determined by the results of MCL in that particular iteration. The $yaw$ angle is calculated using trigonometric methods, with the MCL positions estimated in the current and previous iterations.

\subsection{Results and Discussion}
In order to evaluate the proposed method, a series of experiments have been carried out using the datasets presented in Section \ref{section:experiments_datasets}. Firstly, the performance of the different intermediate layers of the MinkUNeXt network has been evaluated for fine localization. A comparison has also been made between the results obtained with the present method and those obtained in other works of the state-of-the-art in large-scale environments. In addition, results have been included differentiating between indoor and outdoor environments. Finally, to verify the applicability of this method to other datasets, the effectiveness was studied with our own data.

\subsubsection{Evaluation of fine localization with MinkUNeXt layers outputs}
A study has been carried out to analyze the fine localization result with the descriptors obtained with all the layers of the MinkUNeXt network. The output of each layer is a point cloud, wherein each point possesses an associated descriptor. It is important to note that this output point cloud does not maintain the same dimensionality as the input point cloud. The reason for this is that MinkUNeXt is a network that possesses a U-Net architecture, wherein the point cloud undergoes a reduction in dimensionality within the layers belonging to the encoder, followed by a subsequent increase in dimensionality within the layers of the decoder.

\begin{figure*}[t]
\centering
\includegraphics[scale=0.33]{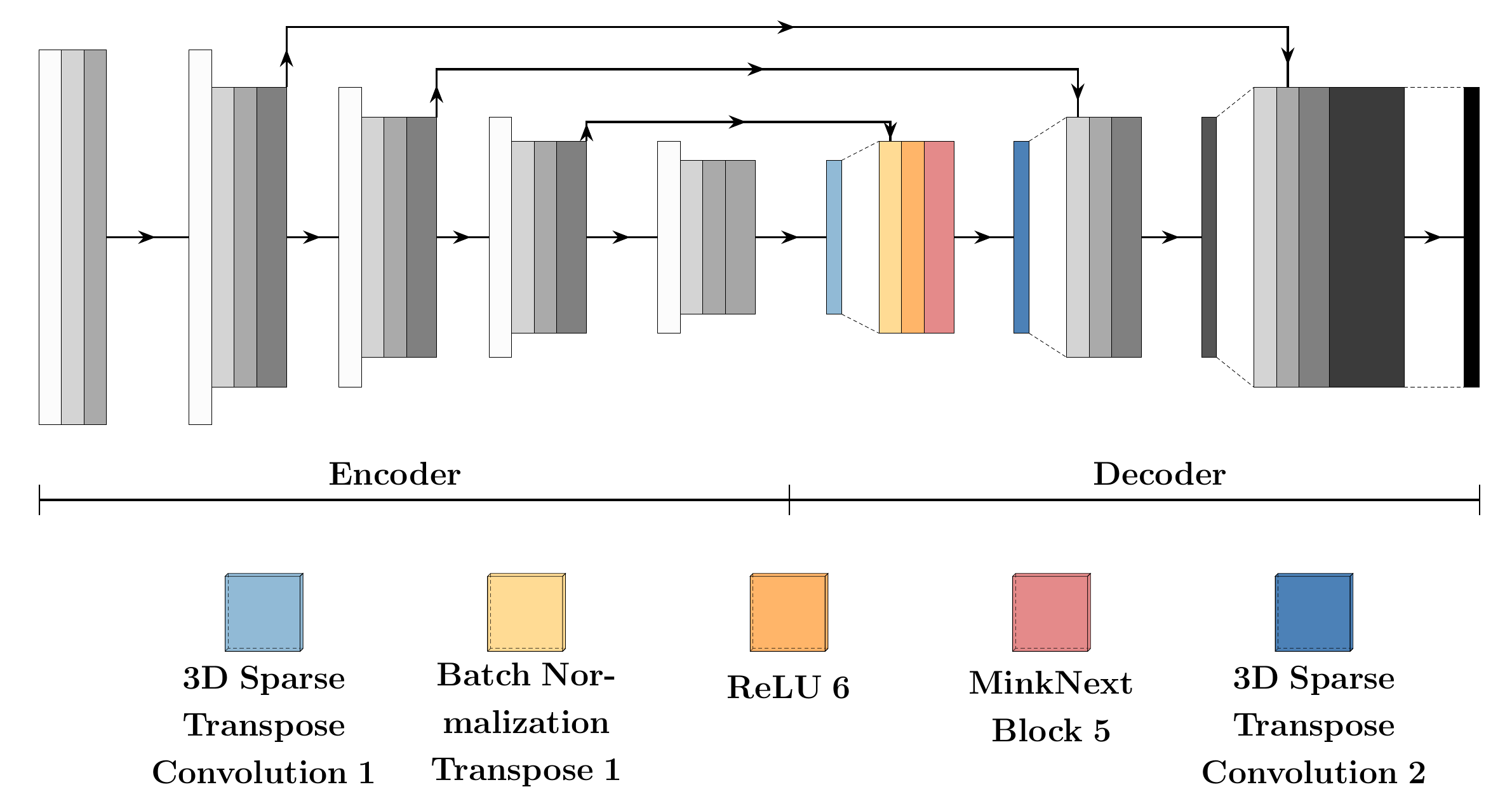}
\caption{MinkUNeXt network architecture. The layers whose output has demonstrated superior performance for the purpose of fine localization are highlighted in colours. The results of these layers are shown in Table \ref{t:results_int_layers}}.
\label{fig:scheme_layers}
\end{figure*}

The results obtained for all the layer outputs were analyzed, with the best results found for the layers located at the beginning of the decoder section of the network. The layers that provide better results are shown in colours in Figure \ref{fig:scheme_layers}. These layers are located immediately after the encoder, at the beginning of the decoder part of the network.
 
The results obtained in these layers are shown in Table \ref{t:results_int_layers}. This table shows that the average error, both in position and orientation, is lower in the 3D Sparse Transpose Convolutional 2. Consequently, this layer will be selected in the method developed for fine localization.

\begin{center}
\begin{table*}
\caption{Mean and median error of the position and orientation obtained with different layers from the MinkUNeXt network in the NCLT dataset. The best results are shown in bold. \label{t:results_int_layers}}
\begin{tabular*}{\textwidth}{@{\extracolsep\fill}lccccc@{}}
    \toprule
       &  & \multicolumn{2}{@{}c}{\textbf{Position}} & \multicolumn{2}{@{}c}{\textbf{Orientation}} 
    \\\cmidrule{3-4}\cmidrule{5-6}\\
     LAYER & Length descriptor & Mean error [m] & Median error [m] & Mean error [deg] & Median error [deg] \\
    \midrule
    3D Sparse Transpose Convolution 1 &192 & 1.72 & 0.97 &5.07 & \textbf{1.83}\\
    Batch Normalization Transpose 1 &192 & 1.79 & \textbf{0.69} & 4.44 & 2.26\\
    ReLU 6 &192 & 1.71 & 0.96 &9.42 &3.20\\
    MinkNext 5 &192 & 3.65 & 1.02 & 11.34 & 3.23\\
    3D Sparse Transpose Convolution  2 &192 & \textbf{1.37} &0.86 & \textbf{3.59} &2.56\\
    \bottomrule
\end{tabular*}
\end{table*}
\end{center}

\subsubsection{Evaluation in large scale environments}
In this section, the performance evaluation of the global localization method is carried out. This evaluation is conducted by comparing the results obtained with MCL-DLF, which employs deep learning, against state-of-the-art outcomes and, concurrently, with MCL-ICP, a classical method based on ICP.

The dataset selected for this evaluation is the NCLT dataset. This dataset has been selected to demonstrate the robustness of the method in large-scale environments with many dynamic elements. This will allow us to test whether the method can be generalized to year-round localization using maps taken in the months of January to March.

In order to compare the results obtained with this dataset, a review of the state-of-the-art was carried out, and the Localising Faster method presented in \cite{sun2020localising} was selected to compare the error results in Tables \ref{t:results_position_nclt} and \ref{t:results_orientation_nclt}. This method has been selected due to its relevance as a state-of-the-art approach for global localization, incorporating deep learning techniques and utilizing a challenging dataset with seasonal variations. This tables show the error values obtained in both position and orientation using both methods. With regard to the aforementioned tables, it should be noted that the state-of-the-art results for each month are obtained with 14,000 to 33,000 tests, whereas in our case they are obtained with 1,800 to 5,000.

\begin{center}
\begin{table*}
\caption{Mean and median error in meters of the position obtained with the global localization method in the NCLT dataset. Including fine localization results obtained with local features and with ICP. The results presented in the State-of-the-Art (SOTA) are also included. The best results are shown in bold. \label{t:results_position_nclt}}
\begin{tabular*}{\textwidth}{@{\extracolsep\fill}lcccccc@{}}
    \toprule
       &\multicolumn{2}{c}{\textbf{SOTA \cite{sun2020localising}}} & \multicolumn{2}{c}{\textbf{MCL-DLF (ours)}} & \multicolumn{2}{c}{\textbf{MCL-ICP}} \\
    \cmidrule{2-3}\cmidrule{4-5}\cmidrule{6-7} 
    MONTHS & median error [m] & mean error [m] & median error [m] & mean error [m] & median error [m] & mean error [m] \\
    \midrule
    February & 1.74 & 8.77  & 0.49 & \textbf{0.85}  & \textbf{0.19} & 1.01  \\
    April    & 1.69 & 2.88  & 0.55 & \textbf{0.73}  & \textbf{0.22} & 0.82 \\
    May      & 2.02 & 15.3  & 0.66 & \textbf{3.93}  & \textbf{0.26} & 4.66 \\
    June     & 1.99 & 11.57 & 0.72 &  \textbf{4.69} & \textbf{0.29} & 4.86  \\ 
    August   & 2.13 & 14.06 & 0.57   & \textbf{0.75}  & \textbf{0.23}    & 1.27  \\  
    October  & 2.14 & 17.33 & 0.52  & \textbf{0.77}   & \textbf{0.22}  & 1.31  \\ 
    November & 3.98 & 17.33 & 0.76 & \textbf{2.23}    & \textbf{0.28}  &  2.32   \\ 
    December & 3.59 & 32.08 & 0.64  & \textbf{2.32}   & \textbf{0.25}  & 2.40   \\ 
    \midrule
    Overall  & 2.18 & 16.55 & 0.58   & \textbf{1.82}     & \textbf{0.23}   & 2.00      \\ 
    \bottomrule
\end{tabular*}
\end{table*}
\end{center}

\begin{center}
\begin{table*}
\caption{Mean and median error in degrees of the orientation obtained with the global localization method in the NCLT dataset. Including fine localization results obtained with local features and with ICP. The results presented in the State-of-the-Art (SOTA) are also included. The best results are shown in bold. \label{t:results_orientation_nclt}}
\begin{tabular*}{\textwidth}{@{\extracolsep\fill}lcccccc@{}}
    \toprule
      &\multicolumn{2}{c}{\textbf{SOTA \cite{sun2020localising}}} & \multicolumn{2}{c}{\textbf{MCL-DLF (ours)}} & \multicolumn{2}{c}{\textbf{MCL-ICP}} \\
    \cmidrule{2-3}\cmidrule{4-5}\cmidrule{6-7}  
     MONTHS & median error [deg] & mean error [deg] & median error [deg] & mean error [deg] & median error [deg] & mean error [deg] \\
     \midrule
    February & 3.25 & 6.19  & 1.62 & \textbf{2.42}  & \textbf{1.23} & 4.09   \\
    April    & 3.36 & 4.43  & 1.81 & \textbf{2.77}  & \textbf{1.46} & 5.14  \\
    May      & 3.34 & 9.50  & 2.35 & \textbf{4.74}  & \textbf{1.61} & 7.91     \\
    June     & 3.17 & 7.96  & 2.32 &  \textbf{4.78} & \textbf{1.58}  & 8.89      \\
    August   & 3.66 & 8.52 & 2.06 &  \textbf{2.92} & \textbf{1.47}  & 4.92     \\
    October  & 3.67 & 11.58 &  1.72 & \textbf{2.80}  &  \textbf{1.29} & 4.55    \\
    November & 5.12 & 17.98 & 2.60    & \textbf{5.14}     & \textbf{1.66}    & 7.62 \\
    December & 4.72 & 14.51 & 2.72  & \textbf{4.13}     & \textbf{1.43}    & 6.45 \\
    \midrule
    Overall  & 3.65 & 4.99  & 1.99    & \textbf{3.39}     & \textbf{1.42}   & 5.61      \\
    \bottomrule
\end{tabular*}
\end{table*}
\end{center}

\vspace{-2cm}
Table \ref{t:results_position_nclt} shows the estimation position results. It can be observed that the position obtained with SOTA has higher values of mean error with respect to its median. This observation suggests the presence of considerable localization errors in certain areas. The use of MCL-DLF reduces these median and mean errors, thereby increasing precision. The use of MCL-ICP has resulted in an improvement to the median error, whereas the mean error has increased slightly compared to the deep local features method. This may be due to ICP being less robust and having more outliers than deep local features approach.

Table \ref{t:results_orientation_nclt} shows orientation errors, where the SOTA presents higher median and mean error values than the other two proposed methods. Moreover, similar to the results obtained for the position estimate in Table \ref{t:results_position_nclt}, MCL-DLF provides lower mean errors than MCL-ICP. Therefore, the error values obtained using MCL-DLF and MCL-ICP show that there has been an improvement in both position and orientation accuracy compared to SOTA.

Furthermore, Tables \ref{t:results_position_nclt} and \ref{t:results_orientation_nclt} show that the SOTA method presents certain variability in the mean errors across different months, while MCL-LDF and MCL-ICP, maintains a relatively small range of variability in the mean error over the different months. This verifies that the proposed method is invariant to seasonal changes. 

As mentioned above, the results obtained with MCL-ICP for both position and orientation are better than those obtained with MCL-DLF in terms of median error, but not in terms of mean error. This is because ICP is sensitive to the initial pose of the point clouds; if the initial misalignment is too large, ICP may fail to converge to the correct alignment or it might get stuck in a local minimum. This is why, despite obtaining a higher median accuracy, in many cases where the initial alignment is very inaccurate, it does not operate adequately. 

\begin{figure*}[h!]
    \centering
    \begin{subfigure}[b]{0.24\textwidth}
        \includegraphics[width=\textwidth]{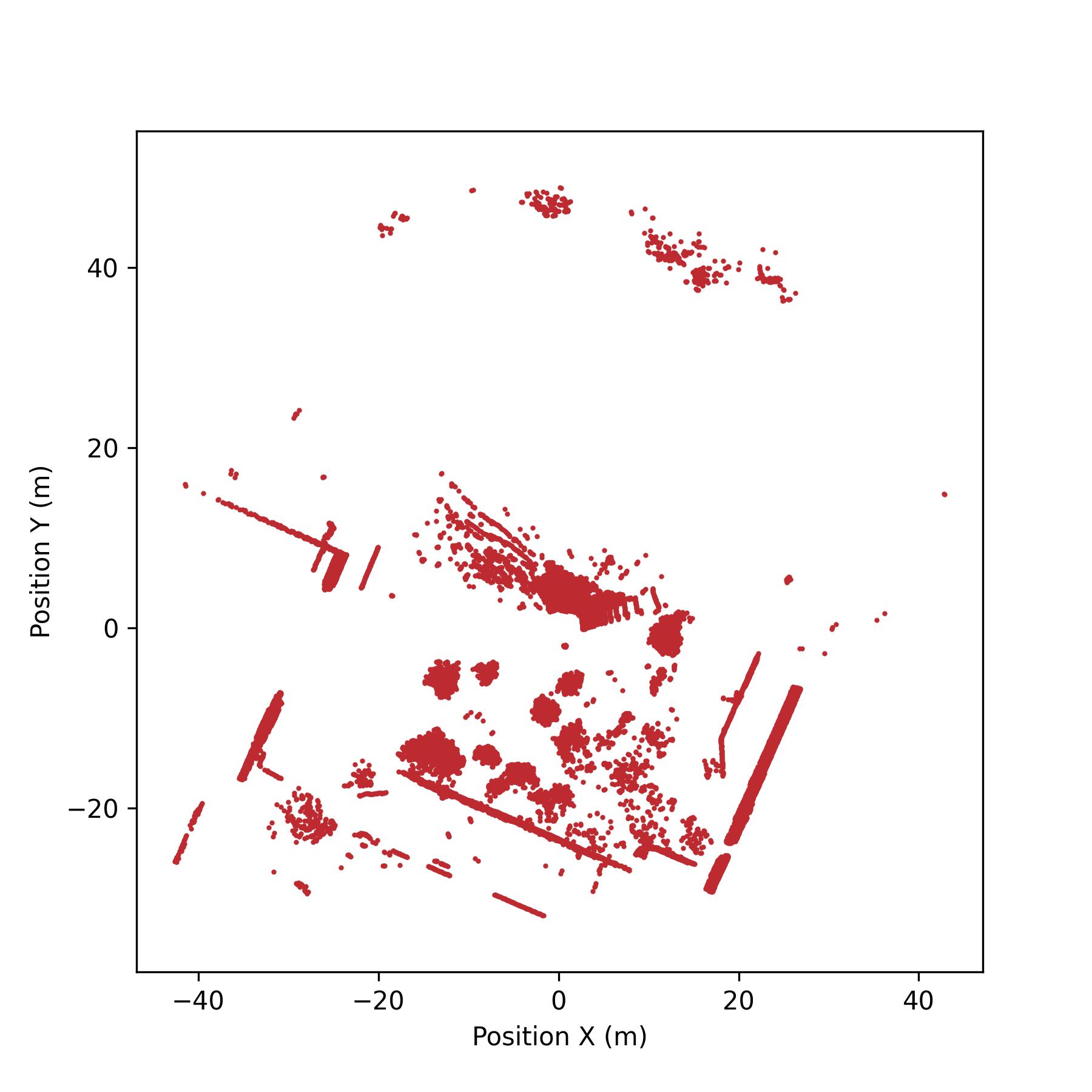}
        \caption{Query point cloud}
        \label{fig:results_global_reg_sub1}
    \end{subfigure}
    \begin{subfigure}[b]{0.24\textwidth}
        \includegraphics[width=\textwidth]{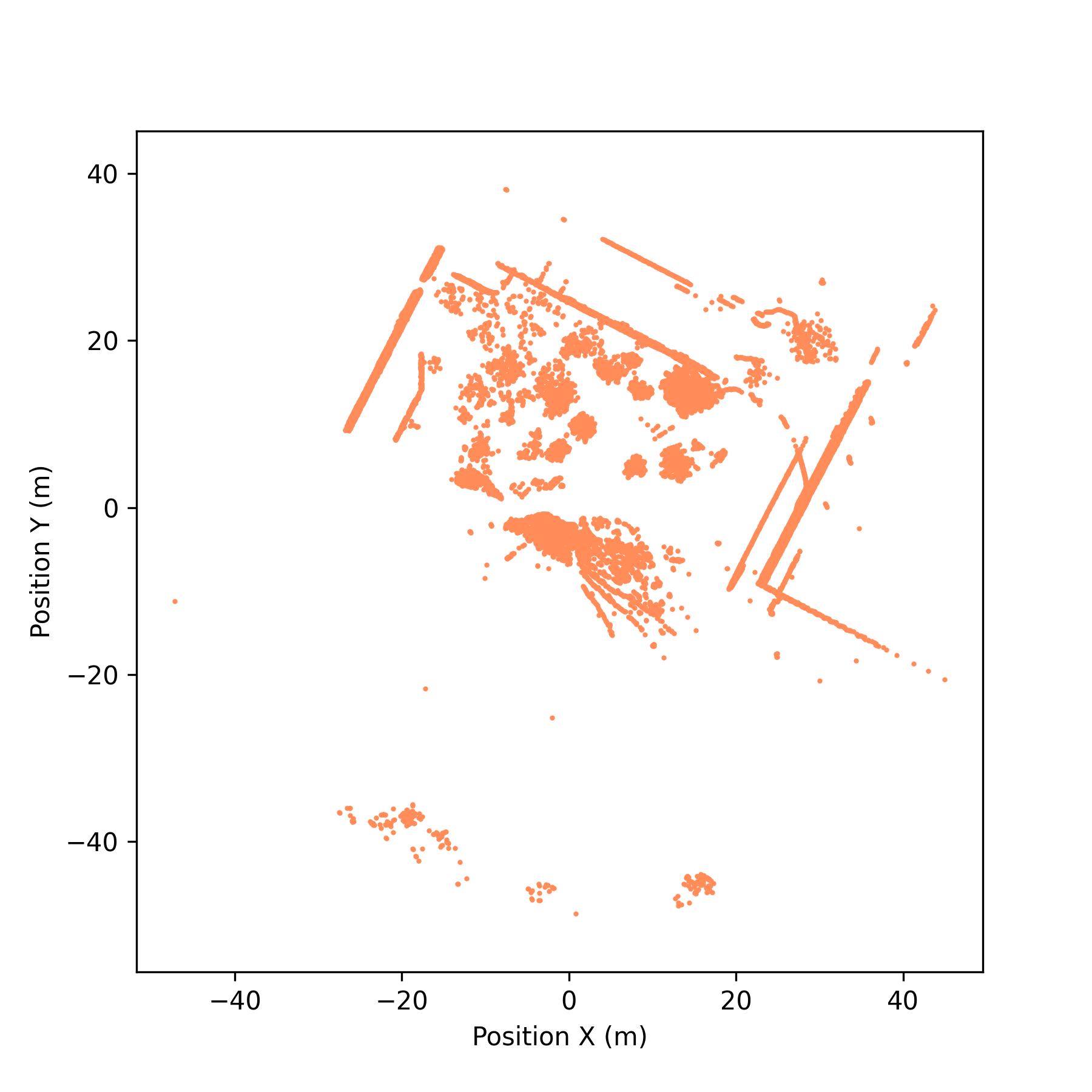}
        \caption{Closest point cloud map}
        \label{fig:results_global_reg_sub2}
    \end{subfigure}
    \begin{subfigure}[b]{0.24\textwidth}
        \includegraphics[width=\textwidth]{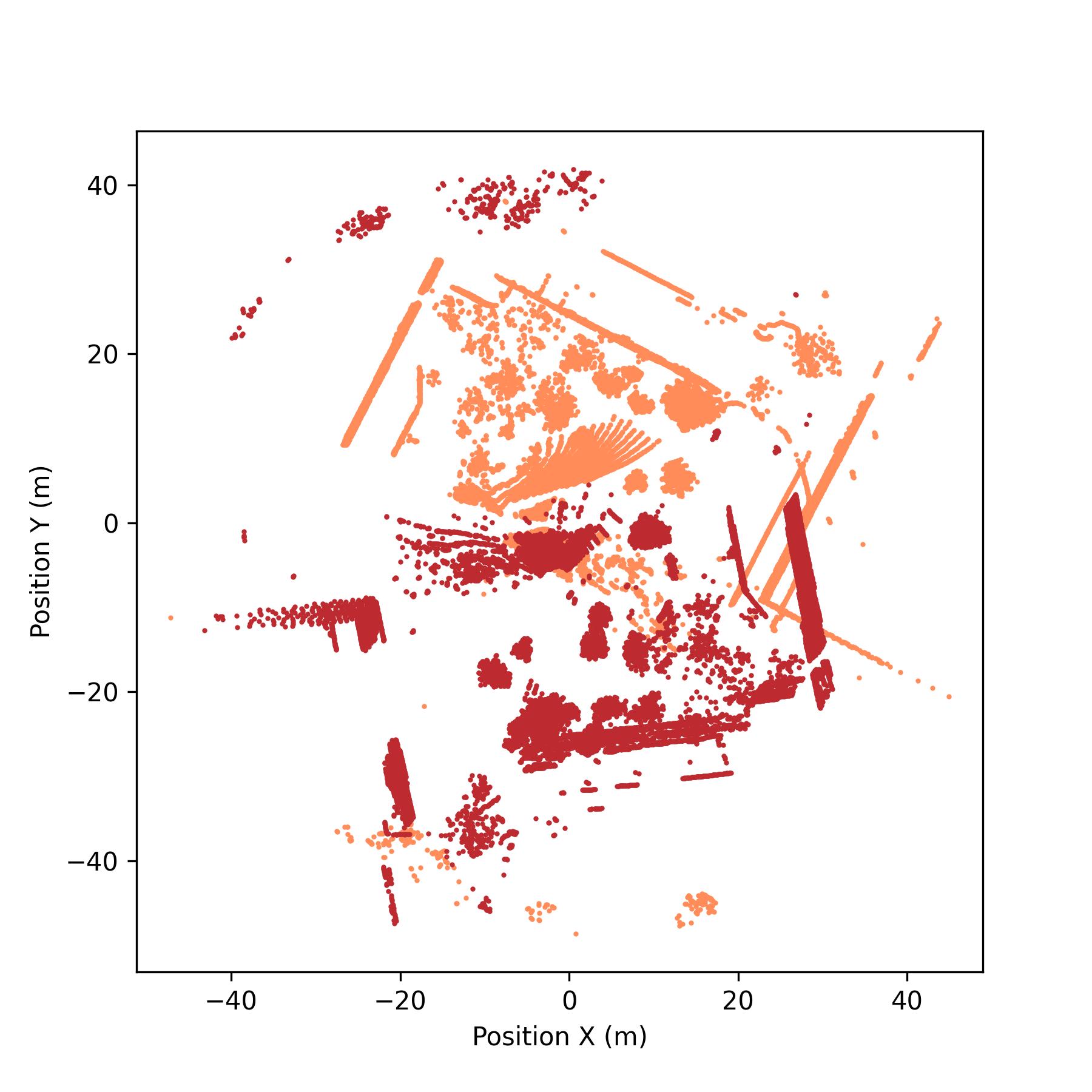}
        \caption{ICP result}
        \label{fig:results_global_reg_sub3}
    \end{subfigure} 
    \begin{subfigure}[b]{0.24\textwidth}
        \includegraphics[width=\textwidth]{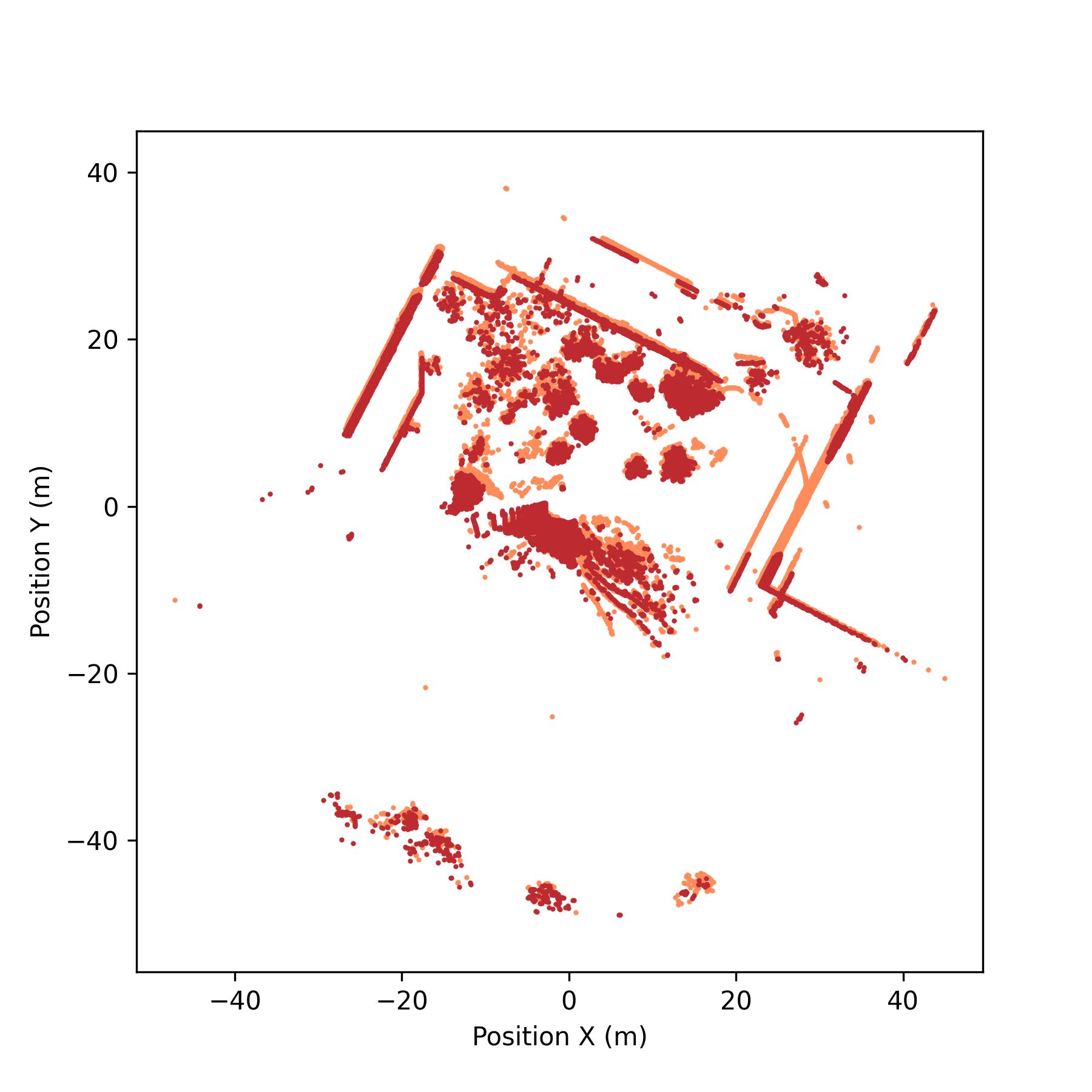}
        \caption{Deep Local Features result}
        \label{fig:results_global_reg_sub4}
    \end{subfigure}
    \caption{Results obtained for the fine localization. Point cloud alignment achieved with the two localization methods evaluated, \textbf{(c)} ICP and \textbf{(d)} Deep Local Features.}
    \label{fig:results_global_reg}
\end{figure*}

On the other hand, the alignment based on deep local features is robust to these problems, as they do not require a prior initial transformation, and rely on the local features of the point clouds to perform the alignment. Figure \ref{fig:results_global_reg} shows an example in which the query cloud captured by the robot and the nearest cloud on the map are initially completely misaligned by an angle of approximately 180 degrees. The ICP result demonstrates a significant difference from the correct alignment. By contrast, with the deep local features, a good point cloud registration result is achieved.

The localization errors in orientation have been compared with other methods that also obtain the robot's orientation, and which have been tested with the NCLT dataset. This comparison has been made with the OREOS \cite{schaupp2019oreos} , OverlapNet \cite{chen2022overlapnet} and DiSCO \cite{9359460} papers described in Section \ref{related_work}. Table \ref{t:comparision_orientation} shows how MCL-DLF outperforms other methods in terms of mean orientation error. It should be noted that the results in Table \ref{t:comparision_orientation} for the other methods do not present point cloud registration. Therefore, it is concluded that the use of fine localization after coarse localization improves the results. Performing values similar to those obtained with DiSCO.

\begin{center}
\begin{table}[h]
\caption{Mean and standard deviation errors in degrees of the orientation estimation in NCLT dataset. The presented results from OREOS, OverlapNet and DiSCO are extracted from \cite{9359460} .\label{t:comparision_orientation}}
\begin{tabular*}{\columnwidth}{@{\extracolsep\fill}lcc@{}}
    \toprule
    Approach in NCLT   &\textbf{Mean [deg]} & \textbf{Std [deg]} \\
     \midrule
    OREOS \cite{schaupp2019oreos}   & 15.95 &  21.31 \\
    OverlapNet \cite{chen2022overlapnet}  & 11.59 &  24.10 \\
    DiSCO \cite{9359460}  & 2.81 &  4.01 \\
    MCL+DLF (ours)      &  3.39 & 5.40  \\
    MCL+ICP     & 5.61 & 12.69     \\
    \bottomrule
\end{tabular*}
\end{table}
\end{center}
\vspace{-0.9cm}
\subsubsection{Evaluation in indoor and outdoor environments}

To test the robustness of the method in indoor and outdoor environments, the results of one of the months were taken and separated into indoor and outdoor environments. Tables \ref{t:results_february_position} and \ref{t:results_february_orientation} show the indoor, outdoor and overall results for the February session.

It can be observed that the errors obtained in both cases are not very different from indoor to outdoor areas. This error variation is greater for MCL-ICP. However, for MCL-DLF, the error is very similar indoors and outdoors, both in position and orientation. Consequently, the MCL-DLF method achieves precise localization regardless of whether the robot is situated in outdoor or indoor environments, demonstrating robustness without dependence on a GPS sensor.
\begin{center}
\begin{table}[h!]
\caption{Mean and median error in meters of the position obtained with the global localization method in the NCLT dataset,  only in February. The results are divided into indoor and outdoor parts of the trajectory, and the results for the entire month are also included. \label{t:results_february_position}}
\begin{tabular*}{\columnwidth}{@{\extracolsep\fill}lcccc@{}}
    \toprule
      & \multicolumn{2}{c}{\textbf{MCL-DLF (ours)}} & \multicolumn{2}{c}{\textbf{MCL-ICP}} \\
    \cmidrule{2-3}\cmidrule{4-5}
      & median [m] & mean [m] & median [m] &  mean [m] \\
      \midrule
    February  & \multirow{2}{*}{0.46} & \multirow{2}{*}{0.48}  & \multirow{2}{*}{0.14} &  \multirow{2}{*}{0.15}  \\
    Indoor & & & & \\
    \midrule
    February & \multirow{2}{*}{0.48} & \multirow{2}{*}{0.87} & \multirow{2}{*}{0.19} & \multirow{2}{*}{1.03}\\ 
    Outdoor & & & & \\
    \midrule
    February  & \multirow{2}{*}{0.48} & \multirow{2}{*}{0.86} &  \multirow{2}{*}{0.19}  & \multirow{2}{*}{1.74}  \\ 
    Overall & & & & \\
    \bottomrule
\end{tabular*}
\end{table}
\end{center}

\begin{center}
\begin{table}[h!]
\caption{Mean and median error in degrees of the orientation obtained with the global localization method in the NCLT dataset. The results are divided in the indoor and outdoor parts of the trajectory. \label{t:results_february_orientation}}
\begin{tabular*}{\columnwidth}{@{\extracolsep\fill}lcccc@{}}
    \toprule
      & \multicolumn{2}{c}{\textbf{MCL-DLF (ours)}} & \multicolumn{2}{c}{\textbf{MCL-ICP}} \\
    \cmidrule{2-3}\cmidrule{4-5}
      & median [deg] & mean [deg] & median [deg] &  mean [deg] \\
      \midrule
    February & \multirow{2}{*}{1.48} & \multirow{2}{*}{1.66}  & \multirow{2}{*}{0.99} & \multirow{2}{*}{1.88}   \\
    Indoor & & & & \\
    \midrule
    February & \multirow{2}{*}{1.63} & \multirow{2}{*}{2.43} & \multirow{2}{*}{1.24} & \multirow{2}{*}{4.14}\\ 
    Outdoor & & & & \\
    \midrule
    February  & \multirow{2}{*}{1.62} & \multirow{2}{*}{2.42} & \multirow{2}{*}{1.23}   & \multirow{2}{*}{4.09}   \\ 
    Overall & & & & \\
    \bottomrule
\end{tabular*}
\end{table}
\end{center}

\subsubsection{Evaluation with in-house data}

In addition, studies have been carried out with our own dataset, collected on the campus of the Miguel Hernandez University, to observe the results for other dataset and to be able to generalize the method presented in other environments.

The peculiarity of this dataset is that it contains many similar environments, as shown in Figure \ref{fig:pcd_map}, so the performance is complex in this case. Thus, the deep local features of the map can be very similar in very distant places. Therefore, an initial coarse localization permits focusing on a single area of the map and hence avoiding mislocalization.

The results of this evaluation are shown in Tables \ref{t:results_position_umh} and \ref{t:results_orientation_umh} where similar accuracy is obtained with respect to the previous results in the NCLT dataset.

In the experiments carried out, it can be seen that the orientation errors obtained with MCL-ICP are much higher than the errors obtained with MCL-DLF. Again, this is because the first method is very sensitive to errors in the initial transformation provided, so if the initial orientation is very different from the correct orientation, the method will converge on an incorrect solution. 

In contrast, MCL-DLF is robust to large changes in orientation, as shown in Figure \ref{fig:results_global_reg}, since MCL-DLF does not rely on an initial transformation to generate the registration, and therefore, is unaffected by a poor initial orientation.

\begin{center}
\begin{table}[h!]
\caption{Mean and median error in meters of the position with the global localization method in the UMH dataset. Including fine localization results obtained with local features and with ICP. The best results are shown in bold. \label{t:results_position_umh}}
\begin{tabular*}{\columnwidth}{@{\extracolsep\fill}lcccc@{}}
    \toprule
    MONTH    & \multicolumn{2}{@{}c}{\textbf{MCL-DLF (ours)}} & \multicolumn{2}{@{}c}{\textbf{MCL-ICP}} 
    \\\cmidrule{2-3}\cmidrule{4-5}
    & median [m] & mean [m] & median [m] & mean [m]\\
    \midrule
    \multirow{2}{*}{April}  &  \multirow{2}{*}{1.62} &   \multirow{2}{*}{1.67} &   \multirow{2}{*}{\textbf{1.40}} &  \multirow{2}{*}{\textbf{1.62}} \\
    & & & & \\
      \multirow{2}{*}{June}  & \multirow{2}{*}{\textbf{1.49}} &  \multirow{2}{*}{1.74} & \multirow{2}{*}{1.64} & \multirow{2}{*}{\textbf{1.70}} \\
      & & & & \\
    \multirow{2}{*}{January}  & \multirow{2}{*}{5.93} &  \multirow{2}{*}{6.76} & \multirow{2}{*}{\textbf{5.80}} & \multirow{2}{*}{\textbf{6.41}} \\
     & & & & \\
    \bottomrule
\end{tabular*}
\end{table}
\end{center}

\begin{center}
\begin{table}[h!]
\caption{Mean and median error in degrees of the orientation with the global localization method in the UMH dataset. Including fine localization results obtained with local features and with ICP. The best results are shown in bold. \label{t:results_orientation_umh}}
\begin{tabular*}{\columnwidth}{@{\extracolsep\fill}lcccc@{}}
    \toprule
    MONTH    & \multicolumn{2}{@{}c}{\textbf{MCL-DLF (ours)}} & \multicolumn{2}{@{}c}{\textbf{MCL-ICP}} 
    \\\cmidrule{2-3}\cmidrule{4-5}
     &  median [deg] & mean [deg] & median [deg] & mean [deg]\\
    \midrule
    \multirow{2}{*}{April}  & \multirow{2}{*}{\textbf{5.76}} &  \multirow{2}{*}{\textbf{9.37}} & \multirow{2}{*}{24.85} & \multirow{2}{*}{20.17} \\
    & & & & \\
    \multirow{2}{*}{June}  & \multirow{2}{*}{\textbf{10.00}} &  \multirow{2}{*}{\textbf{12.81}} & \multirow{2}{*}{24.36} & \multirow{2}{*}{22.17} \\
    & & & & \\
    \multirow{2}{*}{January}  & \multirow{2}{*}{\textbf{8.34}} &  \multirow{2}{*}{\textbf{11.98}} & \multirow{2}{*}{59.59} & \multirow{2}{*}{73.45} \\
     & & & & \\
    \bottomrule
\end{tabular*}
\end{table}
\end{center}

\vspace{-1.5cm}
\section{Conclusions}\label{sec5}
A complete method for accurate global localization from coarse-to-fine has been presented. This approach, denoted as MCL-DLF, involves an initial localization step that roughly estimates the pose, followed by a more accurate refinement step.

The primary benefit of this strategy is improved accuracy.  By first narrowing down the possible locations of the robot in the coarse stage, the subsequent fine localization step can focus its computational resources on a smaller region of the environment.

Deep learning techniques are employed for both stages of the process, enabling the extraction of features from point clouds. This information is then used to compare with features from a pre-existing map, allowing the localization of the robot.

The coarse localization is based on the use of MCL, employing the global point cloud descriptor information as an observation model. This descriptor is extracted from a point cloud using MinkUNeXt. This network employs 3D sparse convolutions, thereby enabling the efficient encoding of extensive three-dimensional geometric environments. The fine localization utilizes point-wise descriptors generated from the intermediate layers of the MinkUNeXt neural network. A comparison has been done between the results provided by the MinkUNeXt-based point cloud registration approach and the classical ICP algorithm. Furthermore, a relevant state-of-the-art method has been compared against the NCLT dataset, as well as other state-of-the-art proposals.

The evaluation of this method on the challenging NCLT dataset and an in-house UMH dataset, shows high accuracy in dynamic environments and under varying environmental conditions. It has been proven that by using a map with data from specific months, precise localization can be achieved in other months with different environmental appearances, without significant variations in the resulting error. Furthermore, the method ensures accurate localization, even in large environments. In addition, the localization method performs well in indoor and outdoor environments. In conclusion, MCL-DLF is a viable option delivering high accuracy in a variety of conditions, including indoor-outdoor scenarios, environmental changes, and the presence of dynamic entities.

\section*{AUTHOR CONTRIBUTIONS}

Conceptualization, M.M., M.B. and D.V.; methodology, M.M.; software, M.M. and A.S.; validation, M.M., M.B., and D.V.; formal analysis and investigation, M.M.; writing—original draft preparation, M.M.; writing—review and editing, M.M., A.S., A.G., M.B. and D.V.; supervision, M.B. and D.V.; project administration, A.G, M.B. and D.V.; funding acquisition, A.G. All authors have read and agreed to the published version of the manuscript.

\section*{ACKNOWLEDGMENTS}
This research work is part of the project TED2021-130901B-I00 funded by MICIU/AEI/10.13039/501100011033 and by the European Union NextGenerationEU/PRTR. It is also part of the project PID2023-149575OB-I00 funded by MICIU/AEI/10.13039/501100011033 and by FEDER, UE.

\section*{CONFLICT OF INTEREST}
The authors declare no conflicts of interest.

\bibliography{wileyNJD-Harvard}

\end{document}